%% file: woftsam.tex
\pdfoutput=1  % arXiv: compile with pdfLaTeX
\documentclass[runningheads]{llncs}

\usepackage{eccv}

\usepackage{eccvabbrv}

\input{preamble}

\combinedtrue  % arXiv: paper + supplementary in one PDF
\ifdefined\combinedbuild\combinedtrue\fi  % the Makefile's `combined` target sets this
\usepackage{hyperref}

\usepackage{orcidlink}

\begin{document}
\title{Segmentation-Guided Homography Estimation \\for Long-Term Planar Tracking}
\titlerunning{Segmentation-Guided Homography Est. for Long-Term Planar Tracking}

% FINAL: Replace with your author list. 
\author{Jonas Serych\orcidlink{0000-0002-8699-9980} \and
Jiri Matas\orcidlink{0000-0003-0863-4844}}

% FINAL: Replace with an abbreviated list of authors.
\authorrunning{J.~Serych, J.~Matas}

% FINAL: Replace with your institution list.
\institute{Visual Recognition Group, Faculty of Electrical Engineering \\
  Czech Technical University in Prague, Czech Republic\\
  \email{\{serycjon,matas\}@fel.cvut.cz}}

\maketitle

\begin{abstract}
  Recent state-of-the-art visual trackers produce high quality and long-term-stable segmentation masks.
  We propose to leverage these strengths for planar object tracking, in which the goal is to estimate a precise 8-degrees-of-freedom homography pose, a geometric representation not estimated by segmentation trackers.
  We present \samh{} -- a planar object tracker that estimates homographies from segmentation mask contours via a training-free pipeline.
  When \samh{} is applied to masks from SAM 2~\cite{ravi2024sam2}, it sets a new state-of-the-art performance on the challenging PlanarTrack~\cite{liu2023planartrack} benchmark by a large margin, +18.4pp on the p@5 metric.
  We further show that segmentation-based and correspondence-based homography estimation are complementary, and propose WOFTSAM, which out-performs all prior methods on both PlanarTrack~\cite{liu2023planartrack} and POT-210~\cite{liang2017planar}.
  We also provide precise re-annotations of PlanarTrack initial poses, enabling more accurate benchmarking in the high-precision p@5 metric.
  The code and the re-annotations are available at \url{https://github.com/serycjon/WOFTSAM}.
\end{abstract}

\section{Introduction}
\label{sec:introduction}
% Planar object tracking is an interesting problem.
Unlike generic visual object tracking, which estimates bounding boxes or segmentation masks, the planar object tracking task requires recovering the full 8-degrees-of-freedom homography transformation~\cite{hartley04thebook} relating the planar target's appearance across frames.
This enables applications in, \eg, augmented reality, movie post-production, and visual servoing, that demand geometric precision beyond mere localization.

% The homography pose has traditionally been estimated using keypoint detection and matching (\eg SIFT~\cite{lowe2004distinctive}), followed by robust model fitting via RANSAC~\cite{fischler1981random}.
Despite the long history of homography estimation~\cite{serych2023planar,zhang2022hvcnet,wang2017gracker,lucas1981iterative,benhimane2007homography,lowe2004distinctive}, planar tracking in general conditions is far from solved.
Challenges include strong perspective distortion, texture-less surfaces, rotation, motion blur, highly reflective surfaces, transparent and virtual targets, and surfaces with dynamically changing appearance as illustrated in~\cref{fig:challenging-examples}.
\input{figures/fig-challenging-results.tex}

While existing correspondence-based approaches perform reliably on well-textured objects and in short-term tracking scenarios, they require sufficiently matchable texture and struggle to recover when the target is temporarily lost due to occlusion, motion blur, or out-of-view motion.
Meanwhile, general object tracking has been transformed by segmentation-based methods such as SAM 2~\cite{ravi2024sam2}, which achieve robust long-term tracking through high-quality mask prediction even on low-textured objects and under strong appearance changes.
However, these methods provide only region-level binary mask output, which lacks the geometric information needed for planar tracking.

This motivates the problem we study: can 8-DoF homographies be reliably extracted from segmentation masks?
This is non-trivial and poses challenges such as occlusion handling and symmetry ambiguities resolution.
For example in the case of quadrilateral targets (used in benchmarks, commonly occuring in reality), any cyclic permutation of corners yields the same quadrilateral, resulting in four homography candidates not distinguishable based on just the mask.

We propose \samh{}, a training-free geometric pipeline that decomposes the mask-to-homography tracking problem into mask boundary line extraction via the Hough transform, identification of the quadrilateral corners, and appearance-based symmetry disambiguation.
While other approaches are conceivable, such as learning a direct masks-to-homography regression, our classical-geometry-based solution requires no task-specific training and serves as a strong baseline.

Applied to masks from SAM 2~\cite{ravi2024sam2}, \samh{} outperforms all prior specialized planar trackers on the challenging PlanarTrack~\cite{liu2023planartrack} benchmark, suggesting that robustness to target appearance changes and temporary tracking failures is the primary bottleneck in current planar tracking.

We further observe that segmentation-based and correspondence-based homography estimation have different strengths depending on the tracking situation: correspondences provide higher precision when texture is visible and tracking is continuous, and during partial occlusions. Segmentation handles blur, re-detection after target loss and texture-less or appearance changing targets.
These situations alternate within individual sequences.
We combine both approaches in WOFTSAM, taking the best of both words (correspondence-based precision and segmentation-based re-detection ability).

The contributions of the paper are:
(i) We formulate the mask-to-homography problem for quadrilateral targets and propose \samh{}, a training-free method that extract homographies from segmentation masks using classical geometry and appearance-based symmetry disambiguation, setting a new state of the art on PlanarTrack~\cite{liu2023planartrack},
(ii) WOFTSAM: a planar tracker that combines correspondence-based precise homographies with robust target re-detection capability of \samh{}, achieving a new state-of-the-art on POT-210~\cite{liang2017planar} and outperforming all prior methods on PlanarTrack~\cite{liu2023planartrack},
(iii) Precise re-annotation of the PlanarTrack~\cite{liu2023planartrack} benchmark initial poses, improving the PlanarTrack suitability for evaluation of high-precision methods.
The effects of the initial frame ground-truth accuracy are significant, see~\Cref{tab:main-results}.

\section{Related Work}
\label{sec:related-work}
% Homography estimation is an old problem. Also in video
% Direct alignment methods (ESM and variants, Lucas Kanade) - find homography that best aligns intensities / edges / features
Planar object tracking via homography estimation is a classical computer vision problem.
Early approaches align an image template to each video frame using intensity-based registration, \eg Lucas–Kanade~\cite{baker2004lucas,lucas1981iterative}.
Efficient Second-order Minimization (ESM)~\cite{benhimane2004real} and its extensions~\cite{chen2019learning,chen2017illumination,lin2019robust} accelerate and improve robustness of Lucas–Kanade by using second-order approximations and feature pyramids.
However, direct alignment methods typically struggle under large blur, occlusion or out-of-view conditions common in real-world planar tracking.

% Keypoints + RANSAC, deep keypoints / descriptors - much more robust in real-world scenarios. Needs lot of texture.
A more robust class of trackers uses feature matching: detect and describe keypoints on the object and match them to each frame.
Standard pipelines use SIFT~\cite{lowe2004distinctive}, SURF~\cite{bay2008speeded} or other classical features and estimate a homography by RANSAC~\cite{fischler1981random}.
Modern variants replace hand-crafted descriptors with learned ones (e.g. GIFT~\cite{liu2019gift}, or LISRD~\cite{pautrat2020online}).
Instead of matching keypoints independently, Gracker~\cite{wang2017gracker} and CGN~\cite{li2023centroid} formulate the task as a graph matching problem.
Current state-of-the-art, WOFT~\cite{serych2023planar} estimates homographies from dense optical flow correspondences.
It pre-warps the current frame using a previous-frame homography estimate so that the residual displacement between the initial frame and the current pre-warped frame is small.
The homography is then estimated by a Weighted Flow Homography (WFH) module, which computes RAFT~\cite{teed2020raft} optical flow between the template and the pre-warped current frame and fits a homography to the resulting correspondences via weighted least squares, where the per-correspondence weights are predicted by a learned module.
% The keypoint-based methods handle occlusion and viewpoint change better, but require rich texture.

% Deep methods HDN (homography decomposition networks), HVC-Net (homography, visibility, confidence), WOFT (pre-warp, homography from optical flow)
Another class of methods is based directly on deep learning.
In HDN~\cite{zhan2021homography}, the homography is estimated by regressing positions of four control points after a rough alignment via an estimated similarity transform.
The HVC-Net~\cite{zhang2022hvcnet} also regresses the four control points and on top of that models uncertainty and occlusions.
PRTrack~\cite{zhang2023multiple} tracks multiple planar targets simultaneously, estimating control points (as peaks of per-point heatmaps) and target segmentation masks. These are then processed together in a multi-stage pipeline, modeling the target motion and occlusions.

% General object tracking currently dominated by segmentation trackers, SAM 2.
% Robust long-term tracking, precise, but doesn't output geometry.
The broader general object tracking field has recently shifted toward segmentation-based methods.
Models like SAM 2~\cite{ravi2024sam2} or DAM4SAM~\cite{videnovic2025distractor} achieve robust long-term tracking with precise segmentation masks.
However, their output (a binary segmentation mask) encodes only the region occupied by the target, without explicit geometric structure.
We propose to utilize the long-term stable segmentation masks for 8-DoF homography estimation.

\section{Method}
\label{sec:method}

Given a video sequence $(I_t)_{t=0}^T$ and a target $X_0$ in the first frame, the goal is to estimate a homography matrix $\mat{H}_t \in \mathbb{R}^{3 \times 3}$ in each frame $I_t$, such that warping the target with the estimated homography gives the target pose $X_t = \mathcal{W}(\mat{H}_t, X_0)$ in the current frame.
The target is usually specified by four control points $X_0 = (x_1, x_2, x_3, x_4), x_i \in \mathbb{R}^2,$ that form a quadrilateral.

\begin{figure*}
  \centering
  \includegraphics[width=1\linewidth]{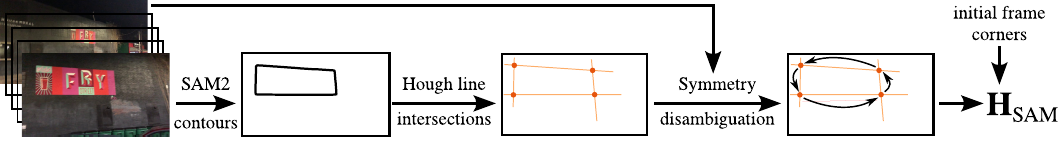}
  \caption{Overview of the \samh{} procedure for estimating homography from a segmentation tracker output.
    First, corners of the SAM 2~\cite{ravi2024sam2} mask are robustly extracted via intersection of Hough lines.
    Next, a symmetry disambiguation process based on a motion model and the target appearance matches the corners with the initial frame and a $\mat{H}_\text{SAM}$ homography is estimated.
    % The \samh{} homography pose provides an initialization for target re-detection ability of the proposed WOFTSAM planar tracker.
  }
  \label{fig:overview}
\end{figure*}

\begin{figure}[t]
  \centering
  \includegraphics[width=\textwidth]{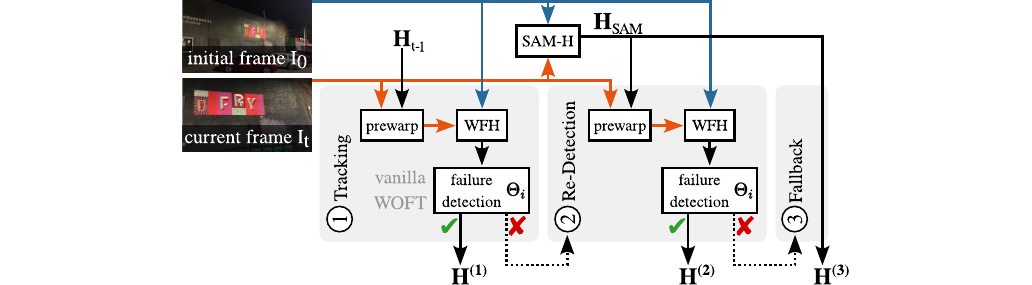}
  \caption{Overview of the proposed WOFTSAM planar tracker.
    Like in WOFT~\cite{serych2023planar}, tracking \circone{} consists of image pre-warping with the previous frame homography $\mat{H}_{t-1}$ and homography estimation via the Weighted Flow Homography (WFH)~\cite{serych2023planar} module.
    If the homography estimation fails (detected by correspondence support set being small), WOFTSAM does a re-detection step \circtwo{}, in which the \samh{} output $\mat{H}_\text{SAM}$ acts as the pre-warping homography.
  If this fails too, WOFTSAM falls back \circthree{} to $\mat{H}_\text{SAM}$.}
  \label{fig:woftsam_overview}
\end{figure}

Recent segmentation trackers\cite{ravi2024sam2,videnovic2025distractor} robustly track the target region throughout most videos in many benchmarks, producing a binary mask~$S_t$ on each frame.
Our goal is to recover the homographies $\mat{H}_t$ from masks $S_t$.
To this end, we propose \textbf{\samh{}} (\cref{fig:overview}), which uses Hough transform to find the edges of the target quadrilateral in the mask boundary, identifies the target corners as intersections of the edges, and resolves the 4-fold symmetry ambiguity (cyclic permutations of the four corners yields the same mask) using a motion model and appearance-based disambiguation.
We then show in \textbf{WOFTSAM} (\cref{fig:woftsam_overview}) that \samh{} homographies, while not pixel-precise, provide a robust re-initialization signal that can be combined with correspondence-based estimation~\cite{serych2023planar} to achieve both high precision and long-term robustness.

\noindent \textbf{\samh{}: From Segmentation Mask to Homography.}
We prompt the SAM~2~\cite{ravi2024sam2} segmentation tracker with a quadrilateral mask specified by the initial coordinates $X_0$ and let it track, providing a mask $S_t$ on each frame.

\emph{Edge and corner extraction.} We fit four lines to the contours of $S_t$ via Hough transform~\cite{duda1972use,hough1962method} and find their intersections $\tilde{X}_t$, keeping only the ones close to the mask and discarding extra intersections.
Detailed description of the Hough transform line search is in the supplementary~\xref{\Cref{sec:more-hough}}{Section~D}.

\emph{Symmetry disambiguation.}
The resulting points cannot be used directly to estimate the homography, since the quadrilateral is invariant to cyclical shifts of the corners due to its symmetrical shape.
The goal of the \emph{symmetry disambiguation} step is to order the detected Hough lines intersections $\tilde{X}_t$ to form $X_t$ which matches to the initial corners $X_0$.
For frame-to-frame corner ordering, we select the cyclic permutation minimizing displacement from the previous frame corner positions $X_{t-1}$.
At typical video frame rates ($\geq$25 fps), inter-frame motion is small relative to the target size, making this a reliable default.

After tracking loss (due to occlusion, out-of-view, or tracking failure), the previous-frame pose is unreliable for disambiguation.
In these cases we use the target appearance to find the best cyclical corner shift.
In particular we compare DINOv2~\cite{oquab2023dinov2} features between the current frame and the initial view warped under each candidate corner ordering.
DINOv2's self-supervised features capture semantic structure robust to geometric imprecisions of the alignment and to significant appearance changes due to reflections, blur, or change of resolution, making them suitable to the task.
% \ttd{TODO: figure showing 4 candidates for which it is hard to find the right orientation}
We require consistent agreement over $\Theta_r = 5$ frames to prevent re-detection on frames where target is partially visible or where the appearance is ambiguous.
After the symmetry disambiguation, a homography is estimated from the matched corners.

\emph{Partial visibility.}
In case all four points are visible and detected, we directly compute the homography $\mat{H}_t$ between the $X_0$ and $X_t$.
In case less than four points are detected, the homography $\mat{H}_t$ is composed of the homography to the previous frame $\mat{H}_{t-1}$ and a residual transformation $\mat{\Delta H}_{t-1 \rightarrow t}$.
When at least two points are visible, $\mat{\Delta H}$ is a 4-DoF similarity transform (translation + rotation + single scale) estimated from the motion of the visible points since the previous frame.
When only a single point is visible, $\mat{\Delta H}$ is a pure 2-DoF translation of the visible point with respect to the previous frame.

\noindent\textbf{WOFTSAM: Precise Tracking With Robust Re-Detection.}
\samh{} homographies $\mat{H}_\text{SAM}$ are robust, but not pixel-precise, since they are derived solely from the mask boundaries.
Correspondence-based methods achieve higher precision when the target is textured and continuously visible, but cannot recover after tracking loss.
WOFTSAM combines both strengths in a simple cascade (\cref{fig:woftsam_overview}).
Given frame $I_t$, WOFTSAM first attempts correspondence-based estimation: the current frame is pre-warped ($\mathcal{W}(\cdot, \cdot)$) with the previous-frame homography $\mat{H}_{t-1}$ and a refined homography is estimated via the Weighted Flow Homography (WFH) module from~\cite{serych2023planar}: \mbox{$\mat{H}^{(1)} = \text{WFH}\left(I_0, \mathcal{W}(\mat{H}_{t-1}^{-1}, I_t)\right)$}.

The reliability of this estimate is assessed by the inlier ratio, \ie, the fraction of flow correspondences consistent with the fitted homography.
If the inlier ratio falls below $\Theta_i = 20\%$, indicating tracking difficulty or loss, WOFTSAM attempts \emph{re-detection}.
The current frame is instead pre-warped with the \samh{} homography $\mat{H}_\text{SAM}$, providing a rough but robust alignment, and WFH is applied again:
\mbox{$\mat{H}^{(2)} = \text{WFH}\left(I_0, \mathcal{W}(\mat{H}_{\text{SAM}_t}^{-1}, I_t)\right)$}.
This leverages \samh{}'s re-detection ability while recovering correspondence-level precision through the flow refinement.
If even this second attempt fails the inlier check, WOFTSAM falls back to the SAM\nobreakdash-H homography $\mat{H}_{\text{SAM}_t}$ directly.
This three-level cascade reflects the observation that correspondence-based estimation should be preferred when it succeeds (higher precision), with \samh{} providing increasingly direct contributions as correspondence quality degrades.

\subsection{Implementation details}
\label{sec:impl-deta}
For the segmentation tracker we use the \texttt{SAM2.1\_hiera\_tiny} variant in default configuration, except for two improvements from DAM4SAM~\cite{videnovic2025distractor} --- we set the memory temporal stride to 5, and disable memory updates when the target is not present.
We found this to perform better than both plain SAM 2~\cite{ravi2024sam2} and the full DAM4SAM~\cite{videnovic2025distractor}.
See~\cref{sec:ablation} for details.
For optical flow homography estimation we use the pre-trained version of RAFT~\cite{teed2020raft} from~\cite{serych2023planar}.

For the symmetry disambiguation, we use a template view either from the initial frame $I_0$, or from a later one where the target appears biggest-so-far, but before any of the four corners becomes lost, \ie, before any re-detections.
This helps in cases when the target has low resolution (far away or zoomed out) on the first frame.
We scale the cropped target views to a standard $224\times 224$ resolution and use the \texttt{dinov2\_vits14\_reg} model~\cite{oquab2023dinov2} features ($16 \times 16$ featuremap resolution, 384 channels).
Each featuremap vector is L2-normalized and the similarity is computed as a dot-product between the warped template view and the current view features.
% For details of the Hough transform line search, see~\Cref{sec:more-hough} in supplementary.

\section{Experiments}
\label{sec:experiments}
We evaluate the proposed methods on two standard planar tracking benchmarks, POT-210~\cite{liang2017planar} and PlanarTrack\textsubscript{TST}~\cite{liu2023planartrack}.

POT-210 tests precision under controlled, isolated challenges (one attribute per video).
PlanarTrack tests robustness under unconstrained conditions with more simultaneous challenges per video, including unconventional targets (transparent, reflective, dynamic appearance).
The two benchmarks demand fundamentally different capabilities, making strong performance on both non-trivial.
% The POT-210 benchmark contains 6+1 videos for each of 30 objects. The six videos focuses on one challenging attribute each: scale change, rotation, perspective distortion, motion blur, occlusions, target out-of-view. The last unconstrained video combines all the above.

% The recent PlanarTrack benchmark is similar to the POT unconstrained category, but more challenging as shown in~\cite{liu2023planartrack}, with extreme scale changes, distractors, reflections, transparent objects and other challenges.

In both benchmarks, the evaluation is based on the \emph{alignment error},
\begin{equation}
  \label{eq:alignment_error}
  e_{\mathrm{AL}}(\mat{H}; \mat{H}^{*}, X) = \sqrt{
    \frac{1}{4} \sum\limits_{i=1}^4
      \left\lVert\mathcal{W}(\mat{H}^{*}, \vect{x}_i) - \mathcal{W}(\mat{H}, \vect{x}_i)\right\rVert^2},
\end{equation}
comparing the positions of four template control points, $X = \{x_1, x_2, x_3, x_4\}$ warped to the current frame by the ground truth homography $\mat{H}^{*}$ and by the estimate $\mat{H}$.

The alignment error is thresholded to produce the \emph{precision} metric.
Following~\cite{serych2023planar} we use two thresholds. The $p@5$ metric, \ie, the fraction of frames where $e_{\mathrm{AL}} < 5\,\mathrm{px}$, and the $p@15$ metric with the $e_{\mathrm{AL}}$ threshold set to 15\,px.

Additionally, we evaluate on the MPOT-3K~\cite{zhang2023multiple} multiple planar target tracking dataset, where the official $e_{\mathrm{AL}}$ threshold is set to 50\,px ($Pr_{\epsilon=50}$ metric, here $p@50$).

\begin{table}
  \centering
  \begin{tabular}{l@{\hskip 5pt}rr@{\hskip 5pt}|@{\hskip 5pt}rr@{\hskip 5pt}|@{\hskip 5pt}ll}
    \toprule
                                                    & \multicolumn{2}{c}{POT-210} & \multicolumn{2}{c}{PT\textsubscript{TST}} & \multicolumn{2}{c}{PT\textsubscript{TST} re-annot}                        \\
    method                                          & p@5         & p@15          & p@5         & p@15                        & p@5                                 & p@15                                \\
    \midrule                                                                      
    SIFT~\cite{lowe2004distinctive,liang2017planar} & 65.8        & 69.6          & 17.6        & 26.7                        & --                                  & --                                  \\
    LISRD~\cite{pautrat2020online,liang2017planar}  & 68.3        & 79.2          & 22.0        & 37.2                        & --                                  & --                                  \\
    HDN~\cite{zhan2021homography}                   & 70.9        & 92.4          & 26.6        & 56.5                        & --                                  & --                                  \\
    CGN~\cite{li2023centroid}                       & 78.5        & 93.3          & --          & --                          & --                                  & --                                  \\
    HVC-Net~\cite{zhang2022hvcnet}                  & \scnd{91.4} & \scnd{95.8}   & 42.0        & 63.1                        & 48.8 \tiny{(\posdiff{+6.8})}        & 64.3 \tiny{(\posdiff{+1.2})}        \\
    WOFT~\cite{serych2023planar}                    & 90.0        & 95.4          & 43.6        & 64.8                        & 48.9 \tiny{(\posdiff{+5.3})}        & 65.9 \tiny{(\posdiff{+1.1})}        \\
    \midrule
    \textbf{WOFTSAM}                                & \frst{91.9} & \frst{97.5}   & \scnd{51.5} & \scnd{77.2}                 & \scnd{57.4} \tiny{(\posdiff{+5.9})} & \scnd{77.6} \tiny{(\posdiff{+0.4})} \\
    \textbf{\samh{}}                                  & 64.4        & 89.0          & \frst{62.0} & \frst{80.0}                 & \frst{62.0} \tiny{(\posdiff{+0.0})} & \frst{80.1} \tiny{(\posdiff{+0.1})} \\
    \bottomrule
  \end{tabular}
  \caption{Precision on the POT-210~\cite{liang2017planar} benchmark with improved ground-truth from~\cite{serych2023planar} and on PlanarTrack\textsubscript{TST}~\cite{liu2023planartrack} (PT\textsubscript{TST}).
    Higher is better.
    The proposed \samh{} tracker sets a new state-of-the-art on PlanarTrack because of its ability to track robustly, including unconventional targets present in parts of the benchmark.
    Thanks to precise correspondence-based homography estimation and the re-detection ability provided by \samh{}, the WOFTSAM sets a new state-of-the-art on POT-210 and out-performs all prior methods on PlanarTrack.
    The last two columns show the effect of precise re-annotation of the initial PT\textsubscript{TST} poses as described in~\Cref{sec:plan-gt-qual}.
    }
  \label{tab:main-results}
\end{table}
\subsection{Main Results}
\label{sec:main-results}
The results of the proposed \samh{} and WOFTSAM planar trackers are summarized in~\cref{tab:main-results}.
Both \samh{} and WOFTSAM substantially outperform all prior methods on PlanarTrack (p@5 +18.4 and +7.9\,pp, p@15 +15.2 and +12.4\,pp over the next-best WOFT~\cite{serych2023planar}).
\begin{figure*}
  \centering
  % (copy-file "~/dev/phd/workarea/flatsam/fig_POT.pdf" "~/dev/phd/doc/papers/cvpr2026/figures/" t)
  % \includegraphics[width=1\linewidth]{fig_POT}\\
  % (copy-file "~/dev/phd/workarea/flatsam/fig_POT_reannot.pdf" "~/dev/phd/doc/papers/cvpr2026/figures/" t)
  \includegraphics[width=1\linewidth]{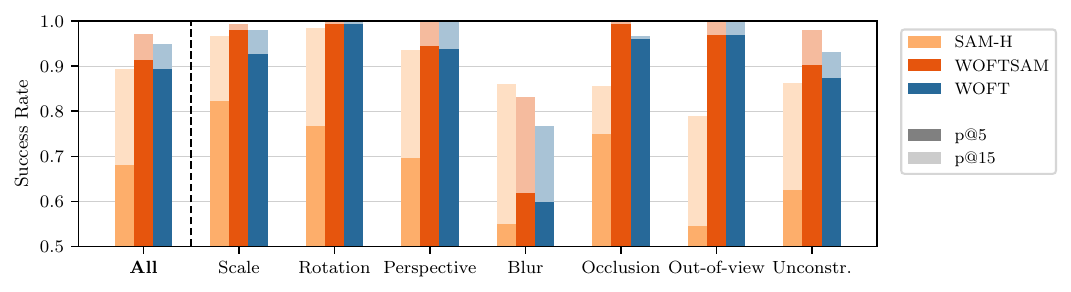}
  \caption{POT-210~\cite{liang2017planar} overall (\emph{All}) and per-attribute results.
    Precision measured on 5\,px and 15\,px alignment error thresholds on the re-annotated GT~\cite{serych2023planar}.
    Although \samh{} is the least precise of the shown methods, using it as a robust re-detection mechanism in the proposed WOFTSAM outperforms the WOFT~\cite{serych2023planar} baseline, almost halving its failure rate on the 15\,px threshold.
    The improvement is particularly high on the sequences where re-detection is needed --- \emph{blur}, \emph{occlusion}, and \emph{unconstrained} --- without decreasing performance on the rest of the benchmark.
  }
  \label{fig:POT-210-challenges}
\end{figure*}

\noindent \textbf{POT-210 results}
On POT-210, where the prior methods already achieve >95\% p@15 and the competition is in the high-precision regime, WOFTSAM sets a new state-of-the-art performance, almost halving the 15\,px alignment errors compared to the baseline WOFT.
The improvement comes from the proposed re-detection capability as confirmed by the challenging-attribute based analysis shown in~\Cref{fig:POT-210-challenges}.
Indeed, the most improved attributes are motion \emph{blur} (target impossible to track due to heavy motion blur, followed by re-detection), \emph{occlusion}, and \emph{unconstrained} which contains both blur and occlusions.
\samh{} mask-based homography estimation by itself is insufficient on POT-210 (64.4 p@5 vs 90.0 for WOFT).
The SAM 2 segmentation masks do not provide boundaries precise enough to find the corners to 5\,px accuracy.
In contrast, the WOFT and WOFTSAM use dense optical-flow-based correspondences from the whole surface of the target, resulting in high precision.
The drop in the p@15 score is caused mainly by the \emph{occlusion} and \emph{out-of-view} attributes, where often only one or two lines of the target quadrilateral are visible.
WOFT and WOFTSAM can infer a precise pose from the texture on the visible part of the target.
Note that once the whole boundary is visible again \samh{} tends to recover.
See~\xref{\Cref{sec:more-pot-results}}{Section~A} in supplementary for success plots and more tracking examples.

\begin{figure}
  \centering
  \begin{minipage}[t]{0.48\linewidth}
    \centering
    % (copy-file "~/dev/phd/workarea/flatsam/PT-timeplot.pdf" "~/dev/phd/doc/papers/cvpr2026/figures/" t)
    \includegraphics[width=1\linewidth]{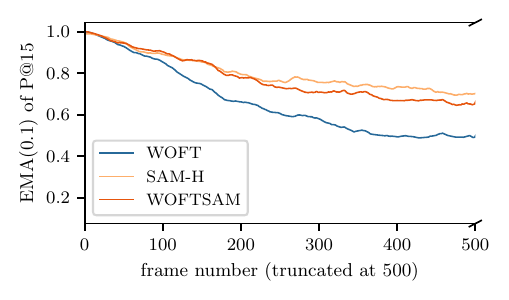}
    \caption{Average p@15 performance on PlanarTrack\textsubscript{TST} as a function of the frame number, smoothed with exponential moving average (EMA).
      After about 3.5 seconds of video ($\approx$ frame 100), the \samh{}-based re-detection starts to boost the WOFTSAM performance compared to the baseline WOFT.
      }
    \label{fig:pt_timeplot}
  \end{minipage}\hfill
  \begin{minipage}[t]{0.48\linewidth}
    \centering
    % (copy-file "~/dev/phd/workarea/flatsam/PT-p15-perseq.pdf" "~/dev/phd/doc/papers/cvpr2026/figures/" t)
    \includegraphics[width=1\linewidth]{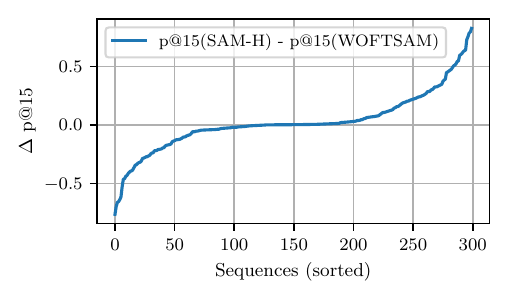}
    \caption{WOFTSAM and \samh{} complementarity on PlanarTrack\textsubscript{TST}~\cite{liu2023planartrack}.
      Per-sequence differences in p@15 scores between \samh{} and WOFTSAM show a balance between the sequences where WOFTSAM excels (\emph{bottom left}) and the sequences where \samh{} excels (\emph{top right}).
    }
    \label{fig:pt-perseq-p15-diff}
  \end{minipage}
\end{figure}

\noindent \textbf{PlanarTrack results.}
On PlanarTrack \samh{} sets a new state of the art performance by a large margin.
This is primarily due to SAM2's ability to robustly re-detect the target after tracking loss and its ability to track unconventional highly reflective or transparent targets, and objects with changing appearance (\eg TV screens), which was not possible with prior planar tracking methods.

WOFTSAM also outperforms all prior methods (+12.4\,pp on p@15), but does not fully match \samh{} (-2.8pp on p@15).
\Cref{fig:pt_timeplot} shows that \samh{} re-detection leads to consistent improvements over the baseline WOFT, with the benefit growing over time as the baseline accumulates irrecoverable failures.

\begin{figure*}
  \centering
  % parallel -j 2 python experiments/paper_sam_vis.py /mnt/tlabdata/serycjon/flatsam/{initfix_woftsam_samfallback_nl,initfix_WACV2026-SAM-H}/ Seq_00341 {} --thickness 6 -ar 1 --crop_margin 0.2 --out_dir /mnt/tlabdata/serycjon/flatsam/figures_supp -w ::: 20 40 60 80
  % (copy-file "~/dev/phd/data/export/flatsam/figures_supp/Seq_00341_0_gt_mask.png" "~/dev/phd/doc/papers/cvpr2026/figures/pt-tracking-examples/" t)
  % (copy-file "~/dev/phd/data/export/flatsam/figures_supp/Seq_00341_20_mask.png" "~/dev/phd/doc/papers/cvpr2026/figures/pt-tracking-examples/" t)
  % (copy-file "~/dev/phd/data/export/flatsam/figures_supp/Seq_00341_40_mask.png" "~/dev/phd/doc/papers/cvpr2026/figures/pt-tracking-examples/" t)
  % (copy-file "~/dev/phd/data/export/flatsam/figures_supp/Seq_00341_60_mask.png" "~/dev/phd/doc/papers/cvpr2026/figures/pt-tracking-examples/" t)
  % (copy-file "~/dev/phd/data/export/flatsam/figures_supp/Seq_00341_80_mask.png" "~/dev/phd/doc/papers/cvpr2026/figures/pt-tracking-examples/" t)
  \includegraphics[width=0.20\linewidth]{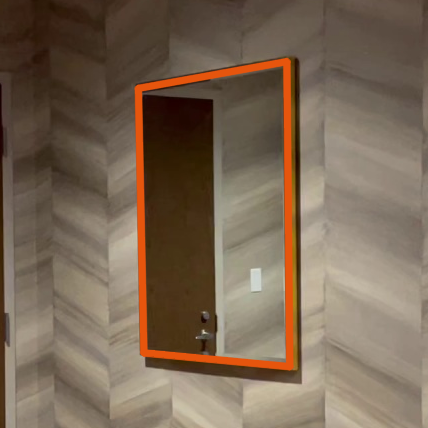}%
  \includegraphics[width=0.20\linewidth]{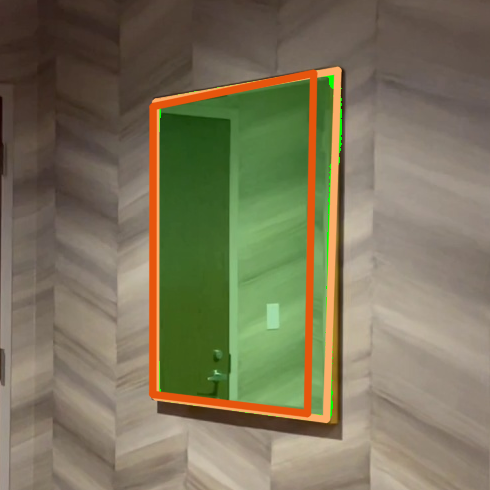}%
  \includegraphics[width=0.20\linewidth]{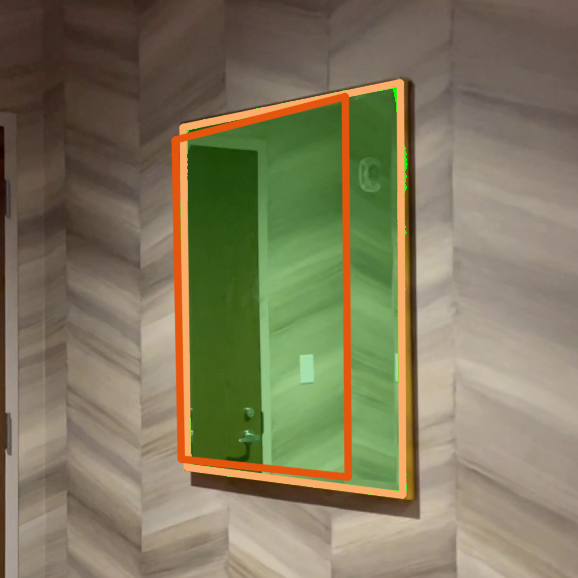}%
  \includegraphics[width=0.20\linewidth]{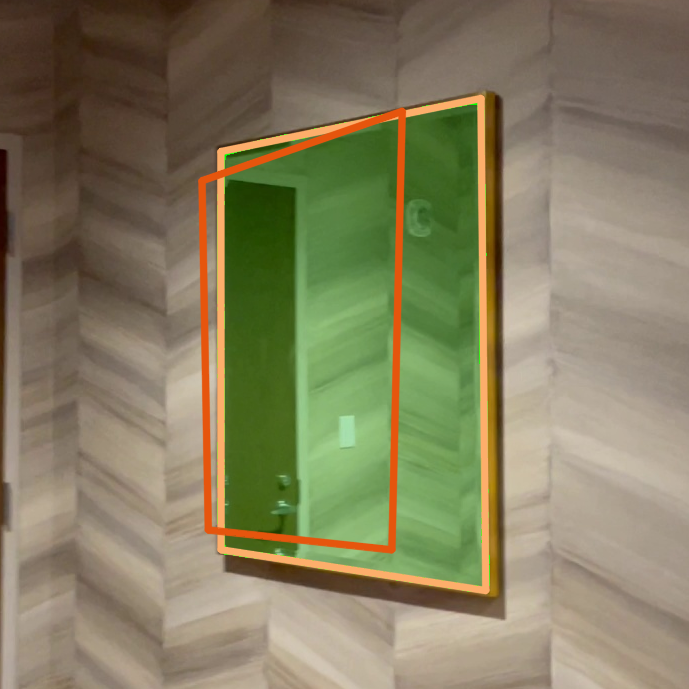}%
  \includegraphics[width=0.20\linewidth]{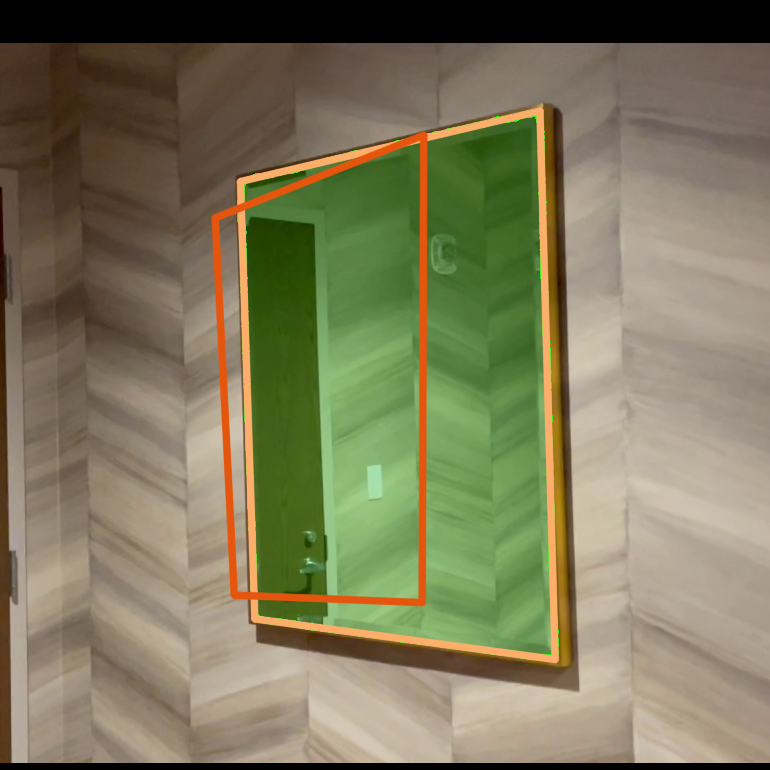} \\
  % parallel -j 2 python experiments/paper_sam_vis.py /mnt/tlabdata/serycjon/flatsam/{initfix_woftsam_samfallback_nl,initfix_WACV2026-SAM-H}/ Seq_00926 {} --thickness 6 -ar 1 --crop_margin 0.2 --out_dir /mnt/tlabdata/serycjon/flatsam/figures_supp -w ::: 20 90 175 195
  % (copy-file "~/dev/phd/data/export/flatsam/figures_supp/Seq_00926_0_gt_mask.png" "~/dev/phd/doc/papers/cvpr2026/figures/pt-tracking-examples/" t)
  % (copy-file "~/dev/phd/data/export/flatsam/figures_supp/Seq_00926_20_mask.png" "~/dev/phd/doc/papers/cvpr2026/figures/pt-tracking-examples/" t)
  % (copy-file "~/dev/phd/data/export/flatsam/figures_supp/Seq_00926_90_mask.png" "~/dev/phd/doc/papers/cvpr2026/figures/pt-tracking-examples/" t)
  % (copy-file "~/dev/phd/data/export/flatsam/figures_supp/Seq_00926_175_mask.png" "~/dev/phd/doc/papers/cvpr2026/figures/pt-tracking-examples/" t)
  % (copy-file "~/dev/phd/data/export/flatsam/figures_supp/Seq_00926_195_mask.png" "~/dev/phd/doc/papers/cvpr2026/figures/pt-tracking-examples/" t) 
  \includegraphics[width=0.20\linewidth]{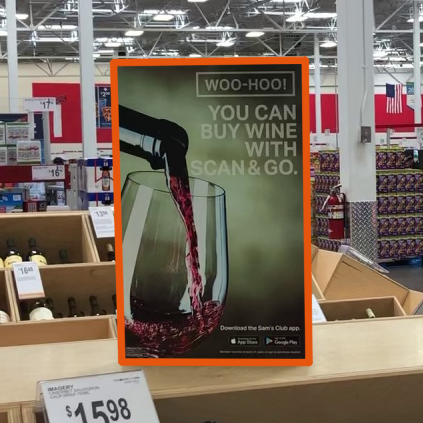}%
  \includegraphics[width=0.20\linewidth]{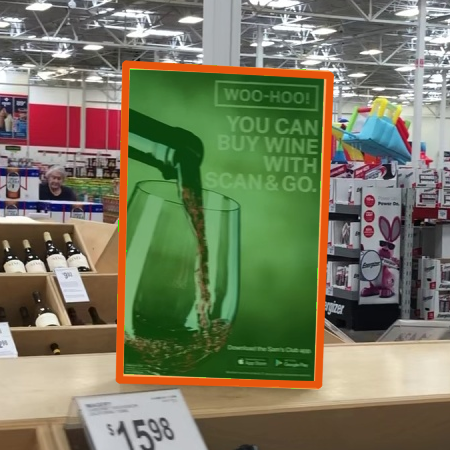}%
  \includegraphics[width=0.20\linewidth]{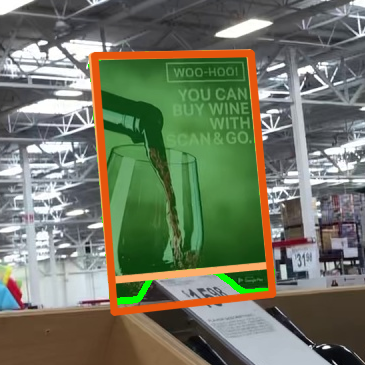}%
  \includegraphics[width=0.20\linewidth]{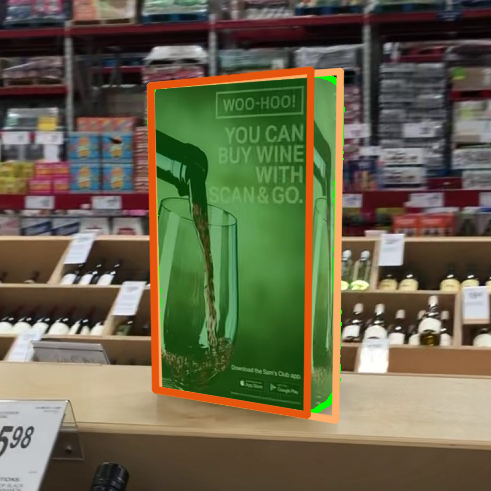}%
  \includegraphics[width=0.20\linewidth]{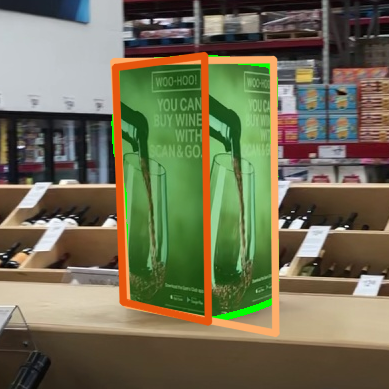}
  \caption{PlanarTrack~\cite{liu2023planartrack} examples, where the correspondence-based WOFTSAM (\textcolor{woftsamcolor}{\emph{dark orange}}) significantly differ from the segmentation-based \samh{} (\textcolor{samhcolor}{\emph{light orange}}).
    SAM 2 mask shown as a \textcolor{green!80!black}{\emph{green}} overlay, frames are cropped around the object.
    \emph{Top:} While SAM and consequently \samh{} tracks the mirror surface as annotated in the dataset,
    WOFTSAM tracks a different planar surface (the doors and part of the adjacent wall) reflected in the mirror.
    % Frames $I_0, I_{20}, I_{40}, I_{60}, I_{80}$ of \texttt{Seq\_00341}. \\
    \emph{Bottom:} Two failure modes of \samh{}.
    (i) SAM 2 segments correctly, but the homography estimation is incorrect due to an occlusion (\emph{middle image}).
    (ii) SAM 2 ``spills'' beyond the prompted surface (\emph{last two images}).
    % Frames $I_0, I_{20}, I_{90}, I_{175}, I_{195}$ of \texttt{Seq\_00926}.
  }
  \label{fig:pt-tracking-examples}
\end{figure*}

Per-sequence analysis (\cref{fig:pt-perseq-p15-diff}) shows that roughly half the sequences are tracked better by \samh{} and the other half by WOFTSAM; a per-sequence oracle selecting between WOFTSAM and \samh{} reaches 86.9 p@15 (+6.9pp w.r.t \samh{}), confirming complementarity.
WOFTSAM excels on well textured objects and during partial occlusions, where dense correspondences provide higher precision than mask boundaries.
\samh{} is stronger on sequences where correspondence-based estimation is unreliable, but this is undetected by the inlier-ratio heuristic, \eg, on a mirror target (see~\Cref{fig:pt-tracking-examples}), where optical flow confidently tracks reflected content that is geometrically consistent, but does not correspond to the intended physical target surface.
Conversely, \samh{} fails when SAM~2 masks do not track the planar object well, \eg, when the planar target is a part of a non-planar object not fully visible in the initial frame and SAM 2 starts tracking the entire object rather than its selected front face.

Developing a more reliable integration of the segmentation-based and the correspondence-based approaches is an open problem.
Nevertheless, even the simple track, re-detect, fallback cascade used in WOFTSAM substantially outperforms all prior methods on both benchmarks.
The large gains from \samh{} demonstrate that segmentation-based homography estimation is a powerful new tool for planar tracking.

\noindent \textbf{MPOT-3K results.} The benchmark~\cite{zhang2023multiple} focuses mainly on multi object tracking (MOT) scenario and consequently uses MOT-style metrics. Instead, we track each object (by object\_id) individually, computing the standard planar tracking metrics.
The results shown in~\cref{tab:mpot3k} demonstrate good performance, especially on the official 50px threshold, on which WOFTSAM outperforms (p@50 95.1 compared to p@50 92.85) the baseline PRTrack~\cite{zhang2023multiple}, which was designed for this particular dataset.
Moreover the PRTrack was evaluated by bi-partite Hungarian matching of the tracker outputs and the GT poses, resulting in over-optimistic results compared to our 1-on-1 prediction-GT evaluation. \Eg, target identity switches do not affect the PRTrack p@50 score.
We have observed that our methods are often visually more precise than the semi-automatically annotated MPOT-3K~GT.
\begin{table}
  \centering
  \begin{tabular}{lrrr}
    \toprule
    method  & p@5  & p@15 & p@50 \\
    \midrule
    WOFTSAM & 67.0 & 89.9 & 95.1 \\
    \samh{}   & 54.2 & 81.2 & 89.9 \\
    \bottomrule
  \end{tabular}
  \caption{MPOT-3K~\cite{zhang2023multiple} evaluation. The p@5 and p@15 results are provided for completeness, the p@50 is more representative due to rough semi-automatic GT annotation of the dataset.}
  \label{tab:mpot3k}
\end{table}

\subsection{Ablations}
\label{sec:ablation}
% \begin{table}
%   \centering
%   \begin{tabular}{lrr}
%     \toprule
%     & \multicolumn{2}{c}{PT\textsubscript{TST}} \\
%     method & p@5 & p@15 \\
%     \midrule
%     SAM 2~\cite{ravi2024sam2} & 59.6 & 77.2 \\
%     DAM4SAM~\cite{videnovic2025distractor} & 58.6 & 77.7 \\
%     \textbf{\tweakedsam} & \frst{62.0} & \frst{80.0} \\
%     \bottomrule
%   \end{tabular}
%   \caption{Ablation --- the choice of the segmentation module for the \samh{} homography estimation.
%     \tweakedsam is the original SAM 2~\cite{ravi2024sam2}, changed to use memory temporal stride 5 and disabling memory updates when the object is not visible.
%   }
%   \label{tab:sam-ablation}
% \end{table}
% \begin{table}
%   \centering
%   \begin{tabular}{rrr}
%     \toprule
%     & \multicolumn{2}{c}{PT\textsubscript{TST}} \\
%     \(\Theta_r\) & p@5 & p@15\\
%     \midrule
%     0 & 58.9 & 76.6\\
%     2 & 61.5 & 79.6\\
%     \frst{5} & 62.0 & 80.0\\
%     8 & 62.0 & 80.1\\
%     \bottomrule
%   \end{tabular}
%   \caption{\samh{} sensitivity to the symmetry disambiguation consistency parameter $\Theta_r$.
%     Disabling the multi-frame consistency requirement ($\Theta_r = 0$) results in a significant performance drop.
%     Results do not improve beyond the default $\Theta_r = 5$.
%   }
%   \label{tab:appearance_run_ablation}
% \end{table}
\begin{table}
  \centering
  \begin{minipage}[t]{0.48\linewidth}
    \centering
    \begin{tabular}{lrr}
      \toprule
      & \multicolumn{2}{c}{PT\textsubscript{TST}} \\
      method & p@5 & p@15 \\
      \midrule
      SAM 2~\cite{ravi2024sam2} & 59.6 & 77.2 \\
      DAM4SAM~\cite{videnovic2025distractor} & 58.6 & 77.7 \\
      \textbf{\tweakedsam} & \frst{62.0} & \frst{80.0} \\
      \bottomrule
    \end{tabular}
    \caption{Ablation --- the choice of the segmentation module for the \samh{} homography estimation.
      \tweakedsam is the original SAM 2~\cite{ravi2024sam2}, changed to use memory temporal stride 5 and disabling memory updates when the object is not visible.
    }
    \label{tab:sam-ablation}
  \end{minipage}\hfill
  \begin{minipage}[t]{0.48\linewidth}
    \centering
    \begin{tabular}{rrr}
      \toprule
      & \multicolumn{2}{c}{PT\textsubscript{TST}} \\
      \(\Theta_r\) & p@5 & p@15\\
      \midrule
      0 & 58.9 & 76.6\\
      2 & 61.5 & 79.6\\
      \frst{5} & 62.0 & 80.0\\
      8 & 62.0 & 80.1\\
      \bottomrule
    \end{tabular}
    \caption{\samh{} sensitivity to the symmetry disambiguation consistency parameter $\Theta_r$.
      Disabling the multi-frame consistency requirement ($\Theta_r = 0$) results in a significant performance drop.
      Results do not improve beyond the default $\Theta_r = 5$.
    }
    \label{tab:appearance_run_ablation}
  \end{minipage}
\end{table}
The~\cref{tab:sam-ablation} supports our choice of the segmentation tracker variant \tweakedsam, \ie the SAM 2~\cite{ravi2024sam2} changed to use memory temporal stride 5 and disabling memory updates when the object is not visible.
These are two simple, but highly effective modifications introduced by DAM4SAM, outperforming both the original SAM~2 and the full DAM4SAM~\cite{videnovic2025distractor}.
The other changes introduced in DAM4SAM are slightly hurting the performance in our mostly distractor-free scenario.
WOFTSAM performance drops only \(-1.1\)\,pp p@5 and \(-1.5\)\,pp p@15 on PlanarTrack\textsubscript{TST} when using plain SAM 2 (as opposed to the \tweakedsam), because the cascade often selects the correspondence-based estimate.

\Cref{tab:appearance_run_ablation} shows tracking results for different values of the $\Theta_r$, which sets the minimal number of consecutive frames with stable appearance-based symmetry disambiguation, after which the appearance-selected cyclical corner shift replaces the one based solely on the motion model.
Not requiring consistence on consecutive frames, \ie, setting $\Theta_r = 0$ results in a -3.4pp p@15 performance drop.
The results do not significantly improve beyond the default $\Theta_r = 5.$

\begin{table}
  \centering
  \begin{tabular}{l@{\hskip 5pt}rr@{\hskip 5pt}|@{\hskip 5pt}rr}
    \toprule
    & \multicolumn{2}{c}{POT-210} & \multicolumn{2}{c}{PT\textsubscript{TST}} \\
    $\Theta_i$                                           & p@5           & p@15                               & p@5         & p@15                                    \\
    \midrule                                                                                             
    10\% & 91.9 & 97.3 & 50.2 & 75.0 \\
    \frst{20\%} & 91.9 & 97.5 & 51.5 & 77.2 \\
    30\% & 91.6 & 97.7 & 52.0 & 77.6 \\
    \bottomrule
  \end{tabular}
  \caption{WOFTSAM sensitivity to the failure detection inlier threshold $\Theta_i$.
    The results are stable over a wide range of values.
    In PT\textsubscript{TST}, the results slightly improve when increasing the threshold, consistently with the WOFTSAM vs \samh{} results.
    Even with the worst-performing 10\% threshold, WOFTSAM out-performs the second-best WOFT by a large margin (+10.2pp).
  }
  \label{tab:inl_thr_ablation}
\end{table}

For failure detection in WOFTSAM, we use the $\Theta_i = 20\%$ inlier threshold from WOFT.
The~\Cref{tab:inl_thr_ablation} contains results with different inlier thresholds, showing that WOFTSAM is stable over wide range of $\Theta_i$ values.

\noindent \textbf{Cascade analysis, component usage and failure rates.}
The frequency of each of the three cascade stages being selected is: \circone{} tracking 85.3\%, \circtwo{} re-detection 3.7\%, \circthree{} fallback 11.0\%.
Re-detection rescue rate is 69.8\%, \ie, the frequency of alignment error of \circtwo{} being at least 15\,px better than that of \circone, and at the same time \circtwo{} is selected. 
The frequency of failure detection miss-classification, \ie, selected \circone{} tracking, but \samh{} would outperform the result by \(>15\)\,px, is only 8.6\%.
In conclusion, most time is spent in the high-accuracy correspondence-based mode and the re-detection mechanisms are useful.

See the supplementary for more ablations (\xref{\cref{sec:more-hough}}{Sec.~D} for Hough transform, \xref{\cref{sec:baseline-experiments}}{Sec.~E} for \samh{} comparison against simple baselines).

\subsection{Computational efficiency}
\label{sec:speed-paper}
We evaluated the WOFTSAM efficiency on the PlanarTrack\textsubscript{TST} dataset.
Overall the proposed WOFTSAM runs at 2.4 FPS using a NVIDIA RTX A5000 GPU, requiring 4.5GB of GPU RAM and Intel(R) Xeon(R) Gold 6326 CPU.
While not real-time, it is enough for applications like tracking in movie post-production.
Like in WOFT (which runs at a similar speed of 2.7 FPS), the most computationally costly part is the optical flow.
The segmentation only takes 42ms per frame, more details in supplementary~\xref{\Cref{sec:speed}}{Section~C}.

% TODO: "The format of table 2,3,4 could be improved."

\subsection{Improving PlanarTrack GT Quality}
\label{sec:plan-gt-qual}
The annotation quality analysis in the recently published journal version of PlanarTrack~\cite{jiao2025planartrack} estimates the mean alignment error of the original ground truth to be 5.71\,px, with more than 10\% of annotations exceeding error of 15\,px.
We have found that inaccurate ground truth accounts for more than half of the p@5 score drop between \samh{} and WOFTSAM as described next.

Most planar trackers --- including those based on optical flow estimation, like WOFT and the proposed WOFTSAM, keypoint-based trackers, and direct alignment methods --- attempt to align the texture of the target in the current frame and in the initial frame.
The resulting homography is then applied to the initial control points to get the current frame control point positions.
In this scenario, the resulting alignment error of the tracker is affected by the ground truth annotation errors on both the initial and the current frame.

\begin{figure}
  \centering
  \begin{minipage}[t]{0.48\linewidth}
  \centering
  % (copy-file "~/dev/phd/workarea/flatsam/PT-init-reannotation-vis-Seq_00027.png" "~/dev/phd/doc/papers/cvpr2026/figures/" t)
  \includegraphics[width=0.49\linewidth]{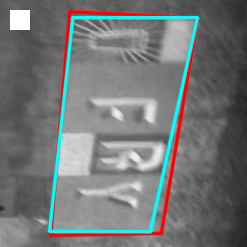}\hspace{2pt}%
  % (copy-file "~/dev/phd/workarea/flatsam/PT-init-reannotation-vis-Seq_00027-382.png" "~/dev/phd/doc/papers/cvpr2026/figures/" t)
  \includegraphics[width=0.49\linewidth]{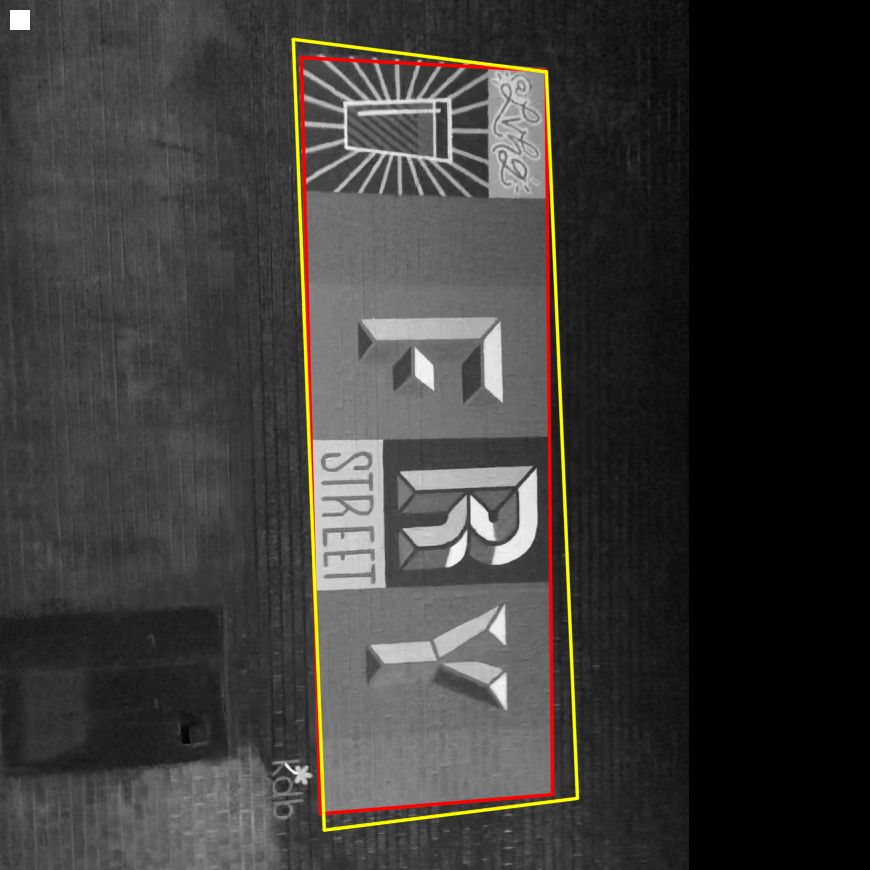}
  \caption{GT annotations (\emph{red}) on frames $I_0$ (\emph{left}) and $I_{382}$ (\emph{right}) of PlanarTrack \texttt{Seq\_00027}.
    Small pixel error on the low-resolution initial target, compared to precise ground truth (\emph{cyan}) results in large error when warped (\emph{yellow}) to a frame with a close-up target view.
    In particular, the corner pixel errors grew from 6.7, 0.5, 11.1, and 4.1 to 20.1, 1.5, 25.2, and 16.8.
    The \emph{top-left} corner of each image shows a $20\,\text{px} \times 20\,\text{px}$ white square serving as a scale indicator.
    Best viewed zoomed in.
  }
  \label{fig:gt-init-scaling}
  \end{minipage}\hfill
  \begin{minipage}[t]{0.48\linewidth}
  \centering
  % (copy-file "~/dev/phd/workarea/flatsam/PT-init-reannotation-vis-Seq_00434.png" "~/dev/phd/doc/papers/cvpr2026/figures/" t)
  % \includegraphics[width=0.49\linewidth]{PT-init-reannotation-vis-Seq_00434}%
  \begin{tikzpicture}[spy using outlines={yellow,magnification=2.5,size=1cm, connect spies}]
    \node[anchor=south west,inner sep=0] (img)  at (0,0) {
      \includegraphics[width=0.49\linewidth]{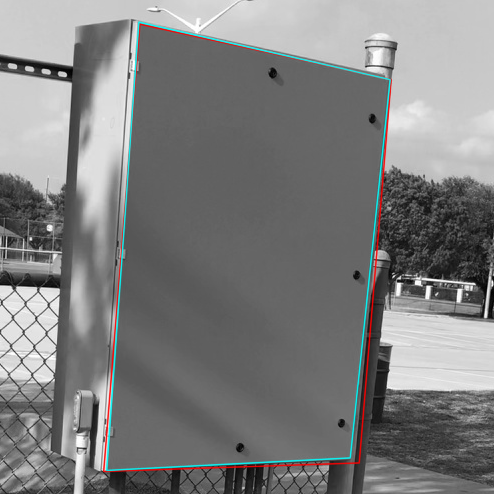}%
    };
    \begin{scope}[x={($ (img.south east) - (img.south west) $ )},y={( $ (img.north west) - (img.south west)$ )}, shift={(img.south west)}]
      % x, y (from bottomleft)
      \node (spy1) at (0.69, 0.09) {};
      \node (spy1to) at (0.5, 0.6) {};
      \spy[magnification=3] on (spy1) in node [left] at (spy1to);
    \end{scope}
  \end{tikzpicture}%
  % (copy-file "~/dev/phd/workarea/flatsam/PT-init-reannotation-vis-Seq_00586.png" "~/dev/phd/doc/papers/cvpr2026/figures/" t)
  % \includegraphics[width=0.49\linewidth]{PT-init-reannotation-vis-Seq_00586}
  \begin{tikzpicture}[spy using outlines={yellow,magnification=2.5,size=1cm, connect spies}]
    \node[anchor=south west,inner sep=0] (img)  at (0,0) {
      \includegraphics[width=0.49\linewidth]{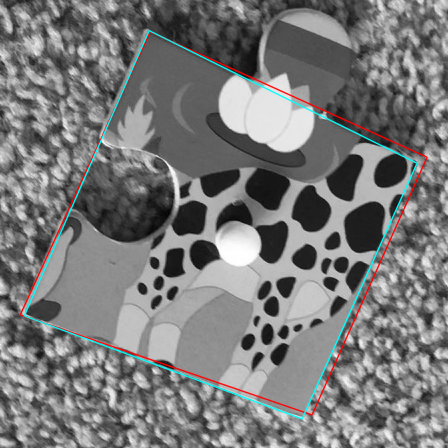}
    };
    \begin{scope}[x={($ (img.south east) - (img.south west) $ )},y={( $ (img.north west) - (img.south west)$ )}, shift={(img.south west)}]
      % x, y (from bottomleft)
      \node (spy1) at (0.67, 0.09) {};
      \node (spy1to) at (0.49, 0.19) {};
      \spy[magnification=3] on (spy1) in node [left] at (spy1to);
      \node (spy2) at (0.90, 0.615) {};
      \node (spy2to) at (0.75, 0.80) {};
      \spy[magnification=2] on (spy2) in node [left] at (spy2to);
    \end{scope}
  \end{tikzpicture}
  \caption{Examples of PlanarTrack\textsubscript{TST} initial frame precise re-annotation.
    The original GT (\emph{red}) goes outside the tracked object, while the new improved GT (\emph{cyan}) exactly aligns with the target.
    Best viewed zoomed in, magnified cutouts in \emph{yellow}.
    The benchmark scores are affected by both the initial and the current frame GT annotation quality for most planar trackers.
    Our re-annotation corrects half of the sources of the benchmarking errors.
    }
  \label{fig:pt-init-reannot}
  \end{minipage}
\end{figure}

On PlanarTrack, the target is often small on the initial frame and increases in size during the sequence (the target is larger than the init in 59\% of frames).
This increases the influence of initial frame GT annotations, \ie a $N$\,px error in the initial frame results in approximately $s_t\cdot N$\,px error in the current frame $I_t$ when the target is $s_t\times$ larger than on $I_0$ as shown in~\Cref{fig:gt-init-scaling}.

We have carefully re-annotated all the PlanarTrack\textsubscript{TST} initial frames with singe pixel or even sub-pixel precision, trying to perfectly align the initial control points with the corners of the target object (see~\Cref{fig:pt-init-reannot}, more details in~\xref{\Cref{sec:re-annot-details}}{Section~B} in supplementary).
The mean alignment error of the original ground truth on initial frames when compared to the precise re-annotation is only $1.9$\,px.
However, this has significant influence on the evaluation of correspondence based trackers due to the aforementioned scaling as shown in~\Cref{tab:main-results}.
In particular, more than half of the PlanarTrack p@5 score difference between WOFTSAM and \samh{} is caused by the inaccurate original initial frame annotations.

Optical flow based methods WOFT~\cite{serych2023planar} and the proposed WOFTSAM greatly benefit from precise initialization.
This is because they propagate any errors in initialization to the whole sequence.
The effect on the \samh{} results are negligible, because \samh{} does not primarily estimate the homography aligning the frames $I_0$ and $I_t$.
Instead, it directly searches for the corners of the tracked object in $I_t$ and the initial corner points only serve as a rough initialization of the SAM 2 tracker.
Note that this also aligns well with what the PlanarTrack annotators did: clicking on the corners of the tracked objects.
In essence, \samh{} is able to ``guess'' the GT even when initialized inaccurately.

Note that we only re-annotated the initial frame of each sequence as improving the annotations on all the 133320 frames would take $\approx 444\times$ more time and the first frame already accounts for approximately half of the benchmarking error.

\section{Limitations and Future Directions}
\label{sec:discussion}
The proposed \samh{} and WOFTSAM methods set a new state-of-the-art on the currently most challenging planar tracking benchmark PlanarTrack.
However, they have some limitations that should be taken into account when deciding what tracker to use for a particular problem.

First, the proposed \samh{} procedure assumes an approximately quadrilateral target shape (which is often the case both in planar tracking benchmarks and in real world) and fails if there are not four stable lines on the target segmentation boundary.
A general mask-to-homography method that would be robust to inaccurate and partially occluded masks and that would work under the strong perspective distortions, rotations and scale changes common in planar tracking is a challenging open problem.

\begin{figure}
  \centering
  % parallel -j 4 python experiments/paper_sam_vis.py /mnt/tlabdata/serycjon/flatsam/{pot_woftsam_samfallback_nl,POT-SAM-H,pot_woft_replica}/ V30_5 {} --thickness 6 -ar 1 --crop_margin 0.2 --dataset pot --bw -w --out_dir /mnt/tlabdata/serycjon/flatsam/figures_supp ::: 341 343 345 347
    % (copy-file "~/dev/phd/data/export/flatsam/figures_supp/V30_5_0_gt_mask.png" "~/dev/phd/doc/papers/cvpr2026/figures/pot-tracking-examples/" t)
    % (copy-file "~/dev/phd/data/export/flatsam/figures_supp/V30_5_341_mask.png" "~/dev/phd/doc/papers/cvpr2026/figures/pot-tracking-examples/" t)
    % (copy-file "~/dev/phd/data/export/flatsam/figures_supp/V30_5_343_mask.png" "~/dev/phd/doc/papers/cvpr2026/figures/pot-tracking-examples/" t)
    % (copy-file "~/dev/phd/data/export/flatsam/figures_supp/V30_5_345_mask.png" "~/dev/phd/doc/papers/cvpr2026/figures/pot-tracking-examples/" t)
    % (copy-file "~/dev/phd/data/export/flatsam/figures_supp/V30_5_347_mask.png" "~/dev/phd/doc/papers/cvpr2026/figures/pot-tracking-examples/" t)
  \includegraphics[width=0.20\linewidth]{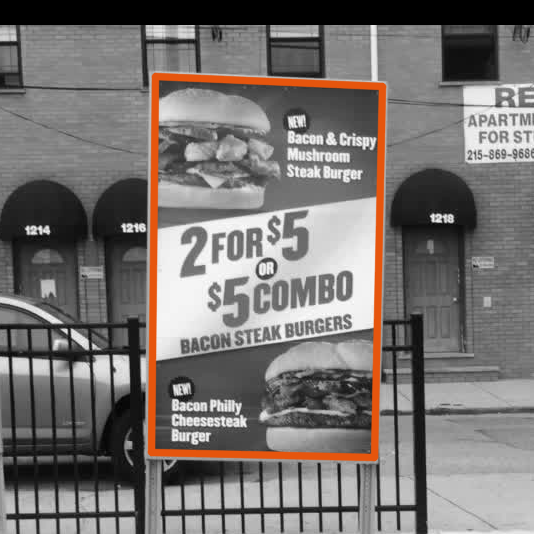}%
  \includegraphics[width=0.20\linewidth]{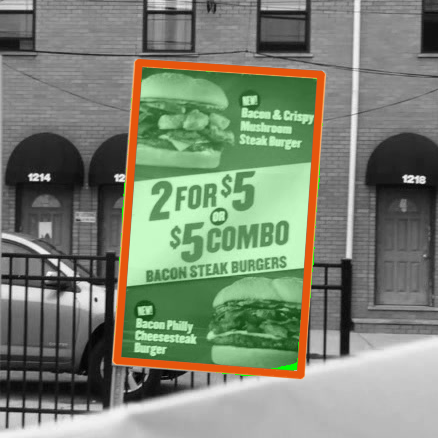}%
  \includegraphics[width=0.20\linewidth]{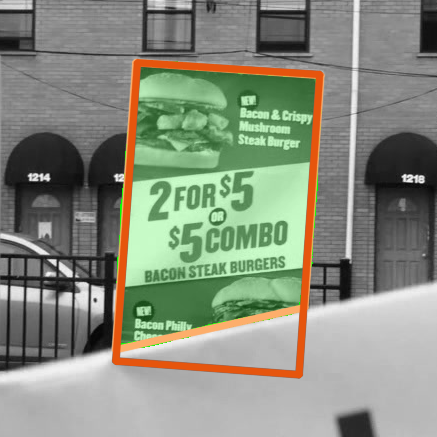}%
  \includegraphics[width=0.20\linewidth]{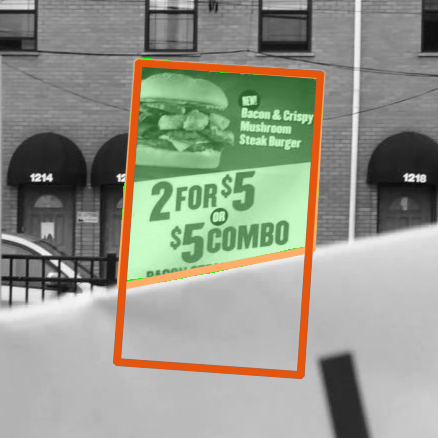}%
  \includegraphics[width=0.20\linewidth]{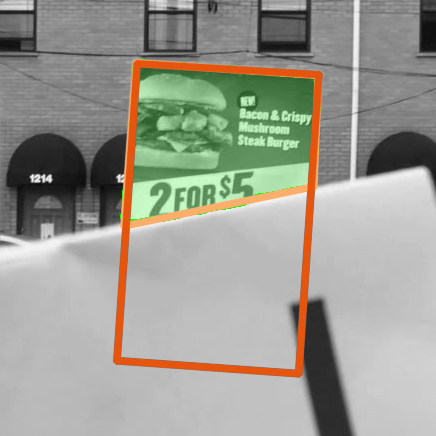}\\
  \caption{An example where \samh{} (\textcolor{samhcolor}{\emph{light orange}}) fails during an occlusion.
    It fits the homography to the visible part of the object (SAM 2 mask shown as a \textcolor{green!80!black}{\emph{green}} overlay).
    The correspondence-based WOFTSAM (\textcolor{woftsamcolor}{\emph{dark orange}}) continues tracking the full object.
    In the POT-210~\cite{liang2017planar} dataset, most occlusions are of this type.
  }
  \label{fig:pot-occlusion}
\end{figure}

Second, there is a room for improvement of the integration of the two paradigms, correspondence- and mask-based, as shown by the per-sequence oracle experiment.
This requires reliably estimating the quality of each homography candidate, which is non-trivial.
Appearance-based criteria fail on texture-less, reflective, or dynamically changing targets, while mask-based criteria have the symmetry ambiguity problem and are unreliable during partial occlusions.
Especially when occluded by objects with linear boundaries that preserve the quadrilateral shape of the segmentation mask (see~\cref{fig:pot-occlusion}).
\input{figures/fig-sam-examples}

Third, the \samh{} procedure is not robust to some types of failures of the underlying segmentation tracker, like when it starts to track the whole 3D object and not just its specified planar face (see~\cref{fig:pt-tracking-examples} and~\cref{fig:sam-examples}).
Also, while \samh{} re-detects well \emph{after} occlusions, it may fail during partial occlusions.
However, the proposed approach is segmentation method agnostic and replacing SAM 2 with a future segmentation tracker (possibly even capable of amodal segmentation behind occlusions) may solve these issues for free.

\section{Conclusions}
\label{sec:conclusions}
We have proposed to leverage modern robust segmentation tracking methods for tracking planar objects with 8-DoF homography poses, a geometrical representation that is not provided directly by the segmentation masks.
The \samh{} tracker sets, by a large margin, a new state-of-the-art on the currently most challenging PlanarTrack~\cite{liu2023planartrack} planar tracking benchmark.
The WOFTSAM tracker combines the strengths of correspondence-based homography estimation with re-detection ability provided by \samh{}, consistently improving over the baseline WOFT~\cite{serych2023planar} and achieving a new state-of-the-art on the POT-210~\cite{liang2017planar} dataset.
The segmentation- and correspondence-based homography estimation have complementary strengths, whose tighter integration and full exploitation is a promising direction for future work.

We also provide precise re-annotations of PlanarTrack initial poses, showing that even small annotation errors significantly affect high-precision evaluation.
Future planar tracking benchmarks should include non-quadrilateral targets and ensure high ground-truth annotation precision to enable meaningful evaluation at tight error thresholds.
We publish the code for \samh{} and WOFTSAM, precise ground-truth re-annotations of the PlanarTrack initial frames, and the annotation tools at~\url{https://github.com/serycjon/WOFTSAM}.

\vspace{5mm}
\noindent \textbf{Acknowledgements.}
This work was supported by the National Recovery Plan project CEDMO 2.0 NPO (MPO 60273/24/21300/21000), the EC Digital Europe Programme project CEDMO 2.0 no.\ 101158609, and by the Research Center for Informatics project CZ.02.1.01/0.0/0.0/16\_019/0000765 funded by OP VVV.

\bibliographystyle{splncs04}
\bibliography{sjo-bib}

% Included only in a combined build (\combinedtrue). For the paper-only
% submission (\combinedfalse) the supplementary is compiled separately via
% supplementary-standalone.tex.
\ifcombined\input{supplementary}\fi

\end{document}

%% file: preamble.tex
\usepackage{microtype}
\usepackage[dvipsnames]{xcolor}
\usepackage{amsmath}
\usepackage{booktabs}
\usepackage{wrapfig}
\usepackage{tikz}

\usetikzlibrary{spy,calc}

\graphicspath{{figures/}}

\newcommand{\mat}[1]{{\mathbf{#1}}}
\newcommand{\vect}[1]{{\mathbf{#1}}}
\newcommand{\bftab}{\fontseries{b}\selectfont}
\newcommand{\frst}[1]{{\bftab{#1}}}
\newcommand{\scnd}[1]{\underline{#1}}
\newcommand{\posdiff}[1]{\textcolor{green!60!black}{#1}}
\newcommand{\tweakedsam}{SAM2$^+$}
\newcommand{\samh}{\mbox{SAM-H}}

\newif\ifcombined
\combinedtrue  % default; overridden by the build-mode switch in the main files
\newcommand{\xref}[2]{\ifcombined#1\else#2\fi}

\usepackage{pifont}
\newcommand{\circone}{\ding{172}}
\newcommand{\circtwo}{\ding{173}}
\newcommand{\circthree}{\ding{174}}

\DeclareMathOperator*{\argmin}{arg\,min}

\definecolor{woftsamcolor}{HTML}{E6550D}
\definecolor{samhcolor}{HTML}{FDAE6B}
\definecolor{woftcolor}{HTML}{276999}

\usepackage[accsupp]{axessibility}  % Improves PDF readability for those with disabilities.

%% file: figures/fig-challenging-results.tex
\begin{center}
  \centering
  \captionsetup{type=figure}
  \begin{subfigure}[t]{0.135\linewidth}
    % (copy-file "/ssh:marr:/mnt/tlabdata/serycjon/flatsam/figures/Seq_00003_0_gt_box.png" "~/dev/phd/doc/papers/cvpr2026/figures/challenging/" t)
    \includegraphics[width=1\linewidth]{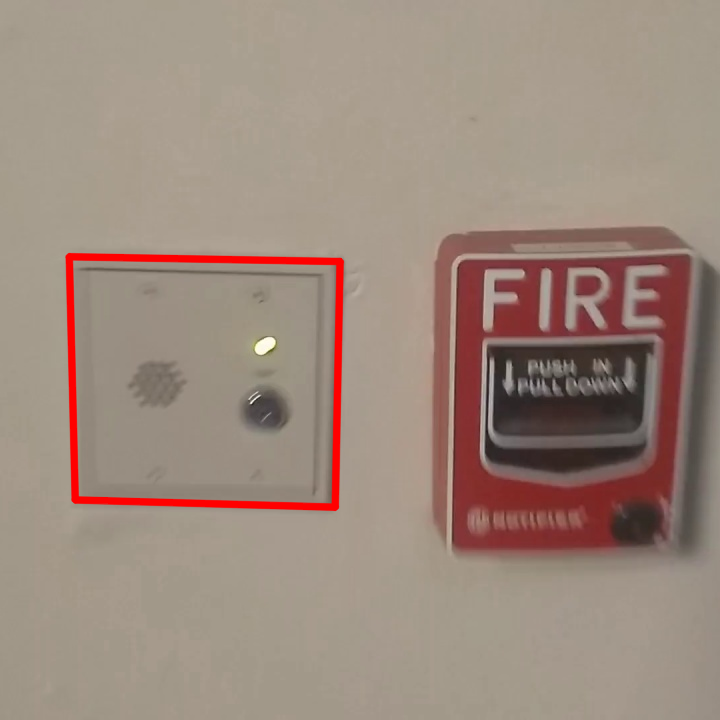}\\
    % (copy-file "/ssh:marr:/mnt/tlabdata/serycjon/flatsam/figures/Seq_00003_23_method_box.png" "~/dev/phd/doc/papers/cvpr2026/figures/challenging/" t)
    \includegraphics[width=1\linewidth]{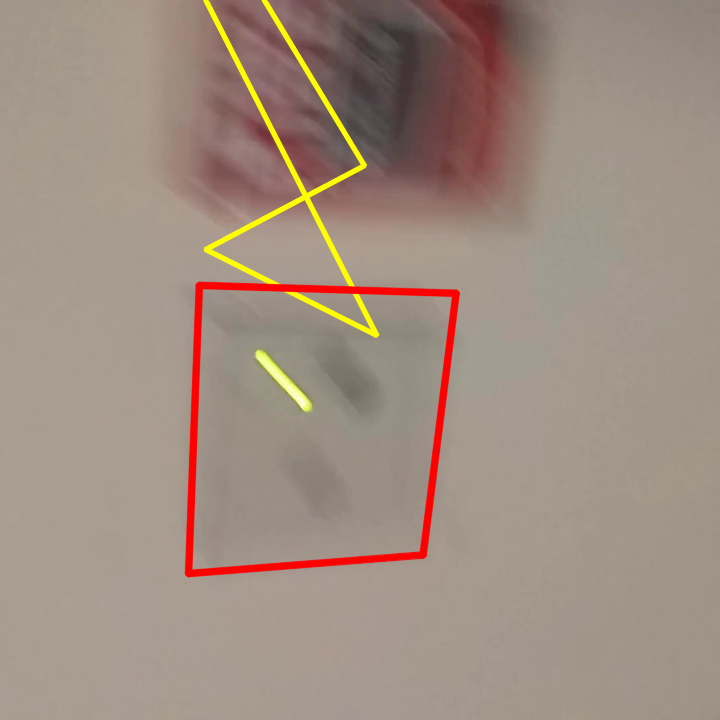}
    \caption{}%\tiny{Motion blur}}
    \label{fig:challenge-blur}
  \end{subfigure}
  \begin{subfigure}[t]{0.135\linewidth}
    % (copy-file "/ssh:marr:/mnt/tlabdata/serycjon/flatsam/figures/Seq_00434_0_gt_box.png" "~/dev/phd/doc/papers/cvpr2026/figures/challenging/" t)
    \includegraphics[width=1\linewidth]{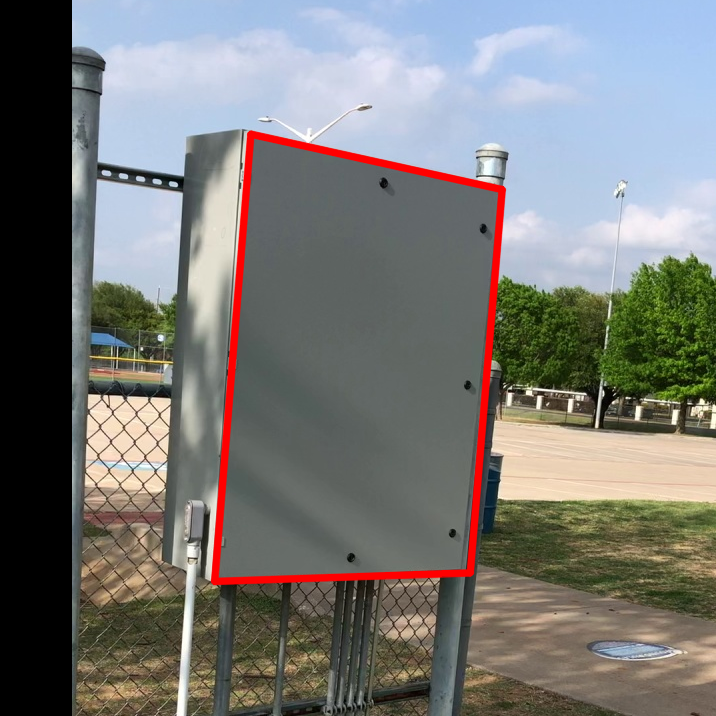}\\
    % (copy-file "/ssh:marr:/mnt/tlabdata/serycjon/flatsam/figures/Seq_00434_499_method_box.png" "~/dev/phd/doc/papers/cvpr2026/figures/challenging/" t)
    \includegraphics[width=1\linewidth]{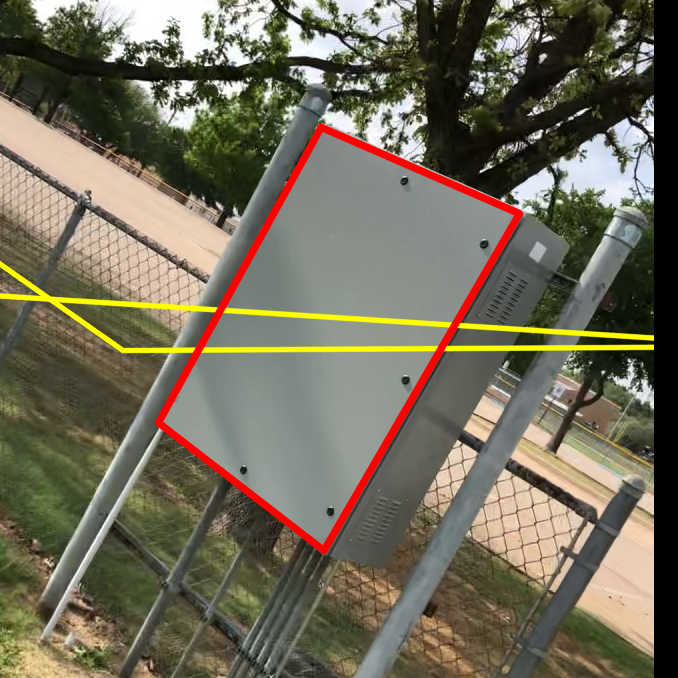}
    \caption{}%\tiny{Textureless}}
    \label{fig:challenge-tless}
  \end{subfigure}
  \begin{subfigure}[t]{0.135\linewidth}
    % (copy-file "/ssh:marr:/mnt/tlabdata/serycjon/flatsam/figures/Seq_00329_0_gt_box.png" "~/dev/phd/doc/papers/cvpr2026/figures/challenging/" t)
    \includegraphics[interpolate=false,width=1\linewidth]{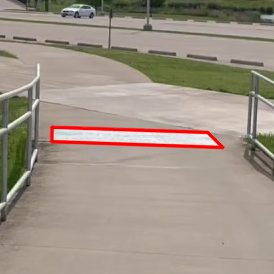}\\
    % (copy-file "/ssh:marr:/mnt/tlabdata/serycjon/flatsam/figures/Seq_00329_255_method_box.png" "~/dev/phd/doc/papers/cvpr2026/figures/challenging/" t)
    \includegraphics[width=1\linewidth]{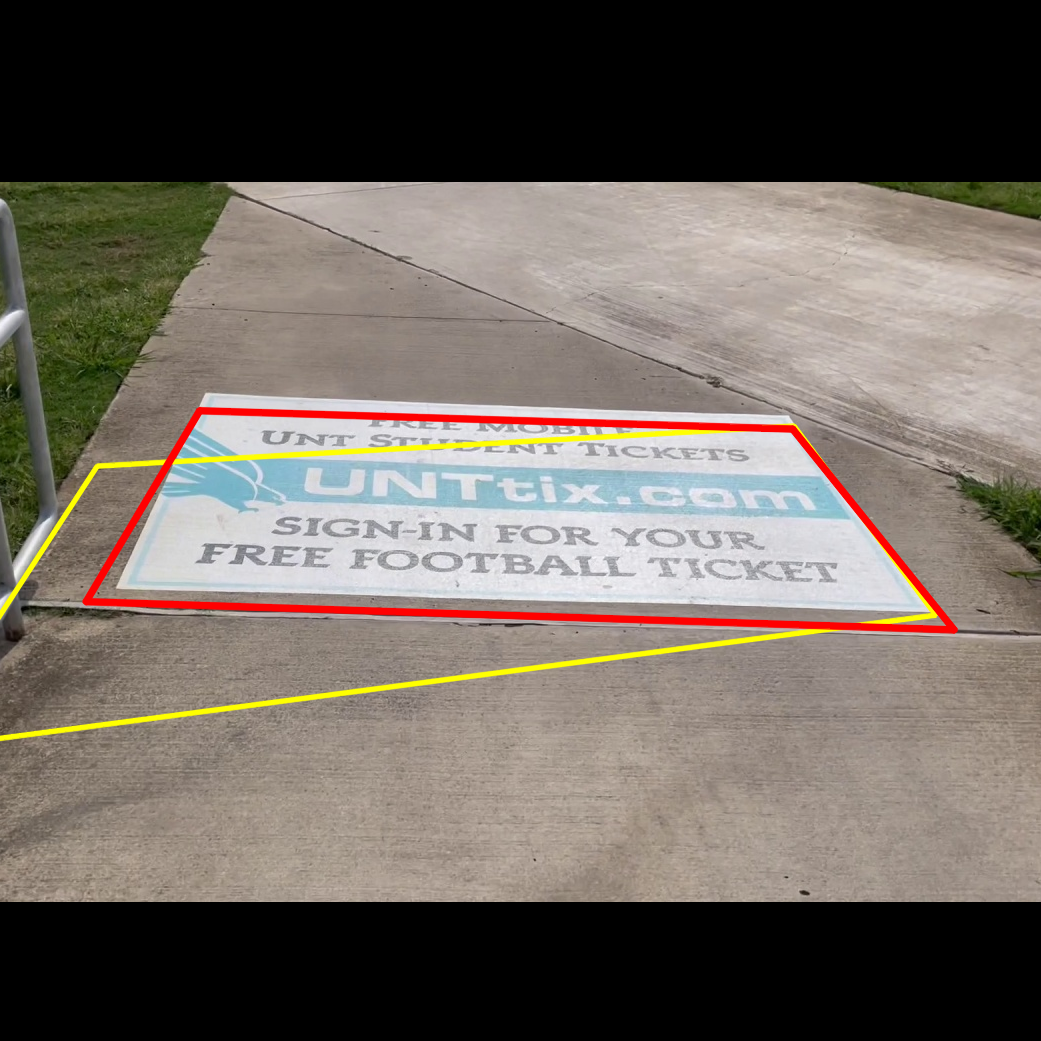}
    \caption{}%\tiny{Zoom $7.5\times$}}
    \label{fig:challenge-zoom}
  \end{subfigure}
  \begin{subfigure}[t]{0.135\linewidth}
    % (copy-file "/ssh:marr:/mnt/tlabdata/serycjon/flatsam/figures/Seq_00319_0_gt_box.png" "~/dev/phd/doc/papers/cvpr2026/figures/challenging/" t)
    \includegraphics[width=1\linewidth]{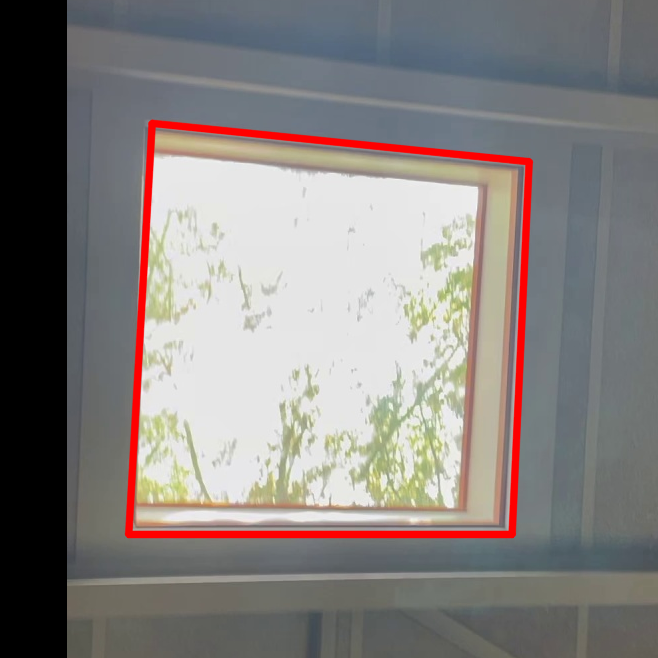}\\
    % (copy-file "/ssh:marr:/mnt/tlabdata/serycjon/flatsam/figures/Seq_00319_491_method_box.png" "~/dev/phd/doc/papers/cvpr2026/figures/challenging/" t)
    \includegraphics[width=1\linewidth]{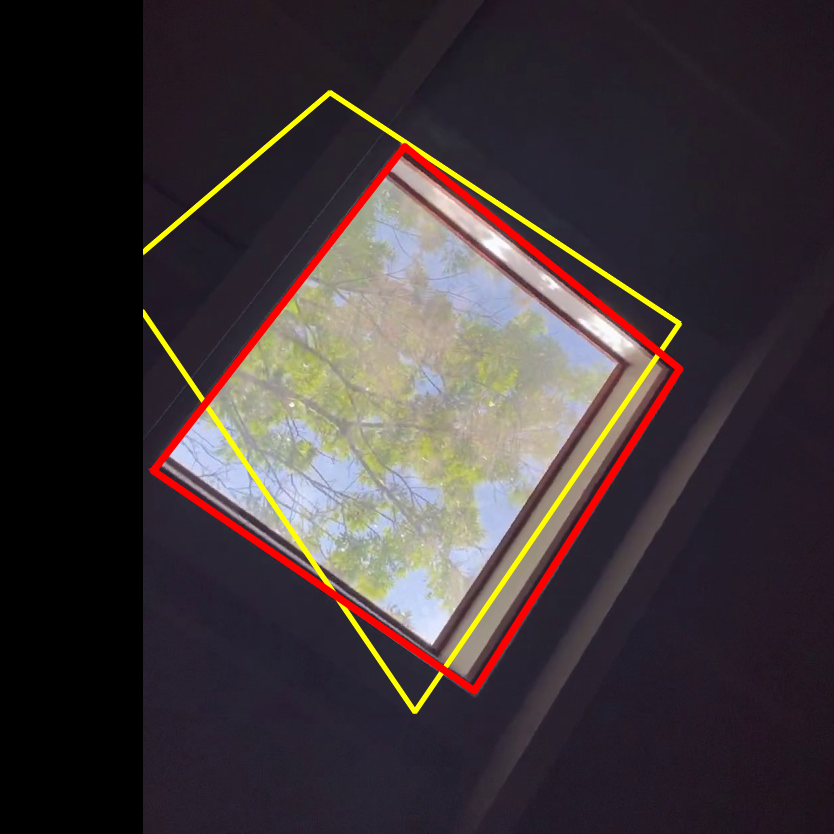}
    \caption{}%\tiny{Virtual plane}}
    \label{fig:challenge-hole}
  \end{subfigure}
  \begin{subfigure}[t]{0.135\linewidth}
    % (copy-file "/ssh:marr:/mnt/tlabdata/serycjon/flatsam/figures/Seq_00341_0_gt_box.png" "~/dev/phd/doc/papers/cvpr2026/figures/challenging/" t)
    \includegraphics[width=1\linewidth]{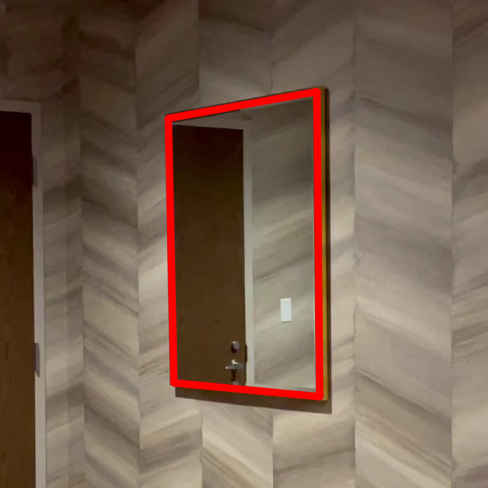}\\
    % (copy-file "/ssh:marr:/mnt/tlabdata/serycjon/flatsam/figures/Seq_00341_342_method_box.png" "~/dev/phd/doc/papers/cvpr2026/figures/challenging/" t)
    \includegraphics[width=1\linewidth]{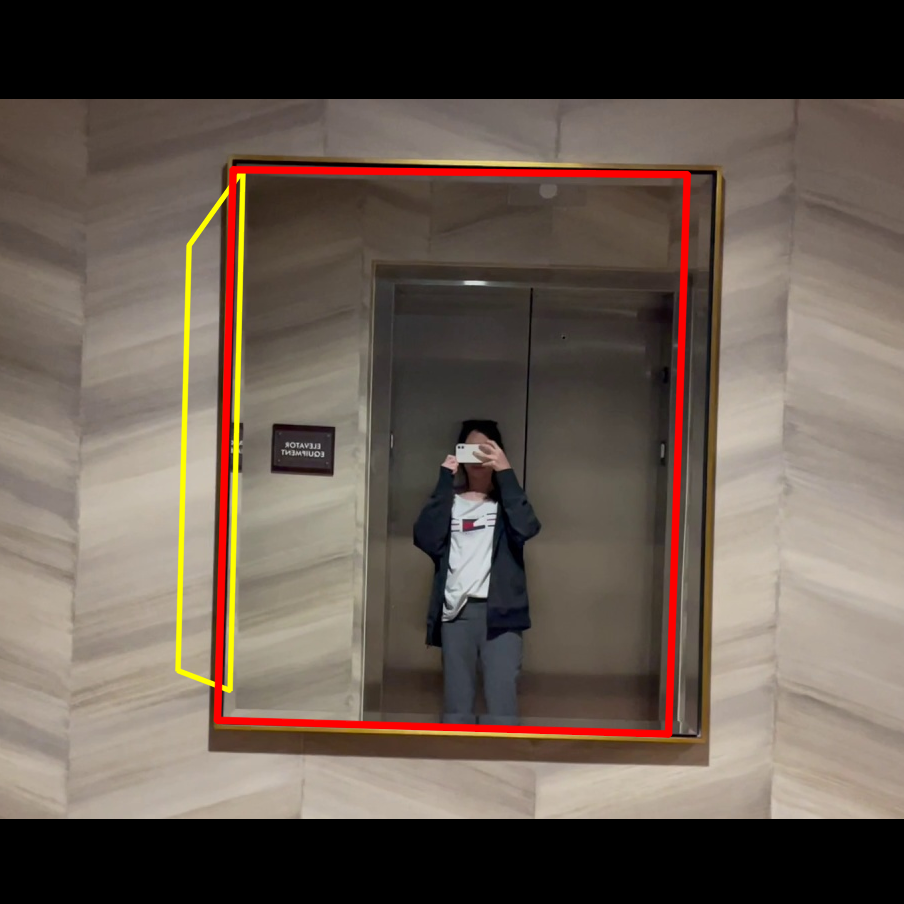}
    \caption{}%Reflective}
    \label{fig:challenge-mirror}
  \end{subfigure}
  \begin{subfigure}[t]{0.135\linewidth}
    % (copy-file "/ssh:marr:/mnt/tlabdata/serycjon/flatsam/figures/Seq_00213_0_gt_box.png" "~/dev/phd/doc/papers/cvpr2026/figures/challenging/" t)
    \includegraphics[interpolate=false,width=1\linewidth]{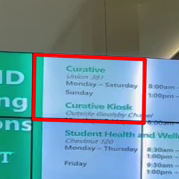}\\
    % (copy-file "/ssh:marr:/mnt/tlabdata/serycjon/flatsam/figures/Seq_00213_472_method_box.png" "~/dev/phd/doc/papers/cvpr2026/figures/challenging/" t)
    \includegraphics[width=1\linewidth]{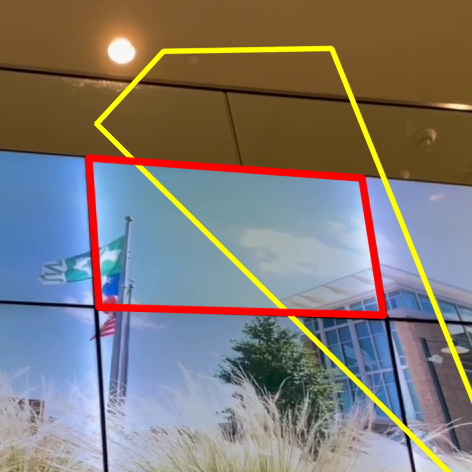}
    \caption{}%Changing content}
    \label{fig:challenge-tv}
  \end{subfigure}
  \begin{subfigure}[t]{0.135\linewidth}
    % (copy-file "/ssh:marr:/mnt/tlabdata/serycjon/flatsam/figures/Seq_00361_0_gt_box.png" "~/dev/phd/doc/papers/cvpr2026/figures/challenging/" t)
    \includegraphics[interpolate=false,width=1\linewidth]{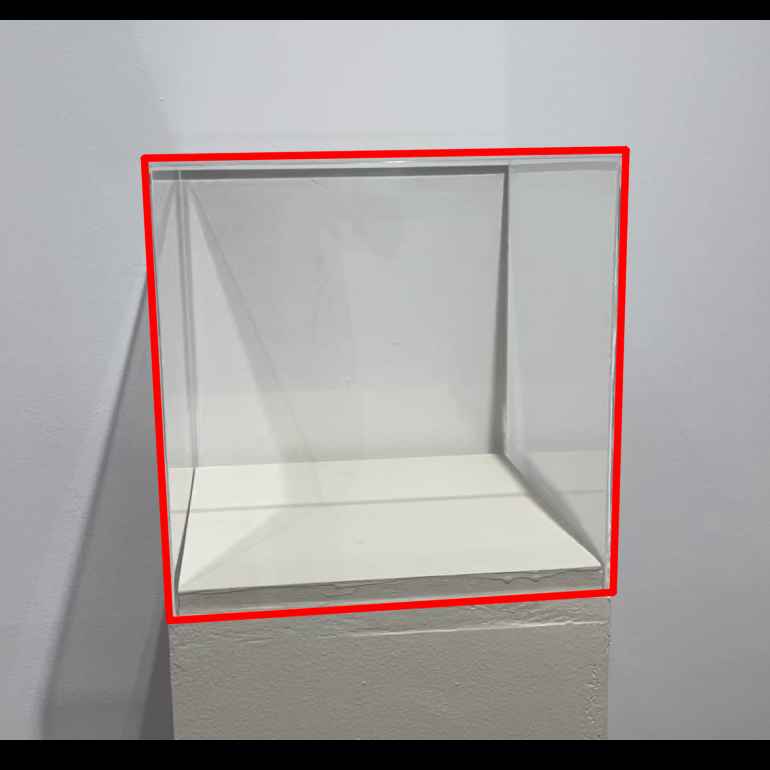}\\
    % (copy-file "/ssh:marr:/mnt/tlabdata/serycjon/flatsam/figures/Seq_00361_455_method_box.png" "~/dev/phd/doc/papers/cvpr2026/figures/challenging/" t)
    \includegraphics[width=1\linewidth]{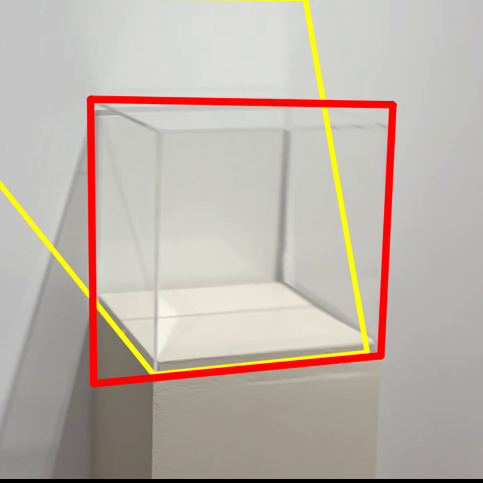}
    \caption{}%Transparent}
    \label{fig:challenge-transparent}
  \end{subfigure}
  \caption{Challenging scenarios from PlanarTrack~\cite{liu2023planartrack}.
    Top: the initial frame, the tracked region in red. Bottom: the proposed \textbf{WOFTSAM} output in red, the  best performing tracker on the dataset in prior art~\cite{liu2023planartrack} WOFT~\cite{serych2023planar} in yellow.
    WOFTSAM out-performs the prior state-of-the-art by introducing a segmentation-based robust re-detection mechanism.
    Examples include planar tracking challenges like perspective distortion, rotation, and scale change \subref{fig:challenge-blur}-\subref{fig:challenge-zoom} as well as unconventional targets that change appearance in the video \subref{fig:challenge-hole}-\subref{fig:challenge-transparent}.
    The last column depicts a partial failure.
}
  \label{fig:challenging-examples}
\end{center}%

%% file: figures/fig-sam-examples.tex
\begin{figure*}[!t]
  \captionsetup[subfigure]{labelfont={tiny},textfont={tiny}} % smaller (a)(b)(c) only for this figure
  \centering
  \begin{subfigure}[t]{0.135\linewidth}
    % (copy-file "/ssh:marr:/mnt/tlabdata/serycjon/flatsam/figures/Seq_00003_0_gt_mask.png" "~/dev/phd/doc/papers/cvpr2026/figures/sam/" t)
    \includegraphics[width=1\linewidth]{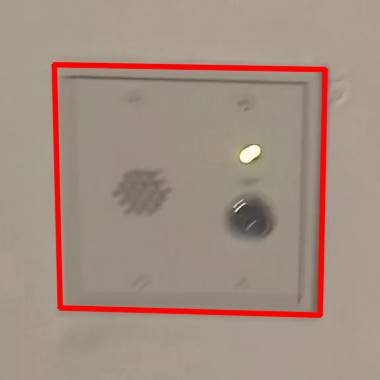}
    % (copy-file "/ssh:marr:/mnt/tlabdata/serycjon/flatsam/figures/Seq_00003_420_mask.png" "~/dev/phd/doc/papers/cvpr2026/figures/sam/" t)
    \includegraphics[width=1\linewidth]{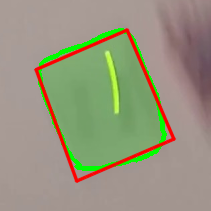}
    % (copy-file "/ssh:marr:/mnt/tlabdata/serycjon/flatsam/figures/Seq_00003_420.jpg" "~/dev/phd/doc/papers/cvpr2026/figures/sam/" t)
    \includegraphics[width=1\linewidth]{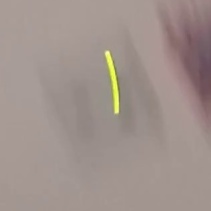}
    \caption{Blur}
    \label{fig:sam-blur}
  \end{subfigure}
  \begin{subfigure}[t]{0.135\linewidth}
    % (copy-file "/ssh:marr:/mnt/tlabdata/serycjon/flatsam/figures/Seq_00212_0_gt_mask.png" "~/dev/phd/doc/papers/cvpr2026/figures/sam/" t)
    \includegraphics[width=1\linewidth]{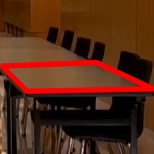}
    % (copy-file "/ssh:marr:/mnt/tlabdata/serycjon/flatsam/figures/Seq_00212_276_mask.png" "~/dev/phd/doc/papers/cvpr2026/figures/sam/" t)
    \includegraphics[width=1\linewidth]{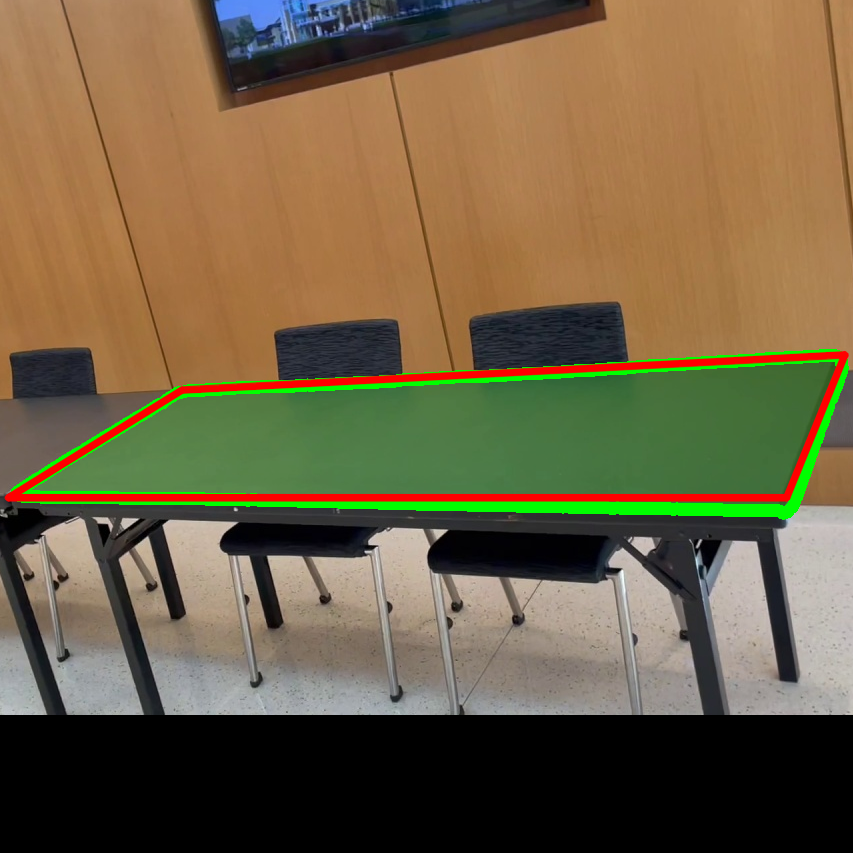}
    % (copy-file "/ssh:marr:/mnt/tlabdata/serycjon/flatsam/figures/Seq_00212_276.jpg" "~/dev/phd/doc/papers/cvpr2026/figures/sam/" t)
    \includegraphics[width=1\linewidth]{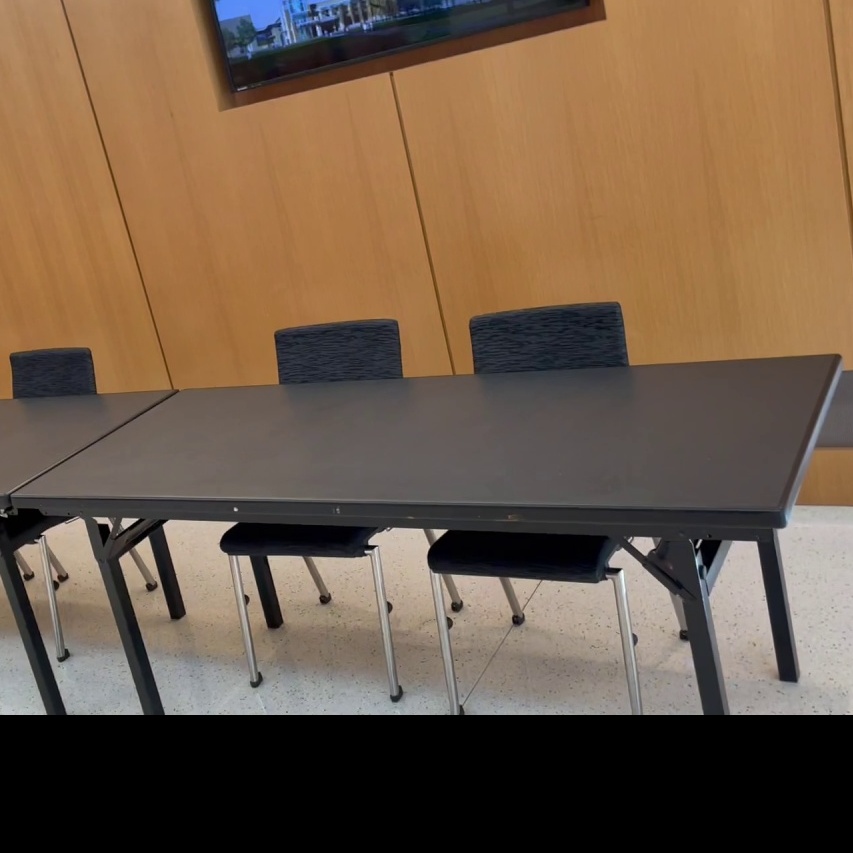}
    \caption{Texture-less}
    \label{fig:sam-tless}
  \end{subfigure}
  \begin{subfigure}[t]{0.135\linewidth}
    % (copy-file "/ssh:marr:/mnt/tlabdata/serycjon/flatsam/figures/Seq_00349_0_gt_mask.png" "~/dev/phd/doc/papers/cvpr2026/figures/sam/" t)
    \includegraphics[width=1\linewidth]{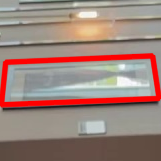}
    % (copy-file "/ssh:marr:/mnt/tlabdata/serycjon/flatsam/figures/Seq_00349_450_mask.png" "~/dev/phd/doc/papers/cvpr2026/figures/sam/" t)
    \includegraphics[width=1\linewidth]{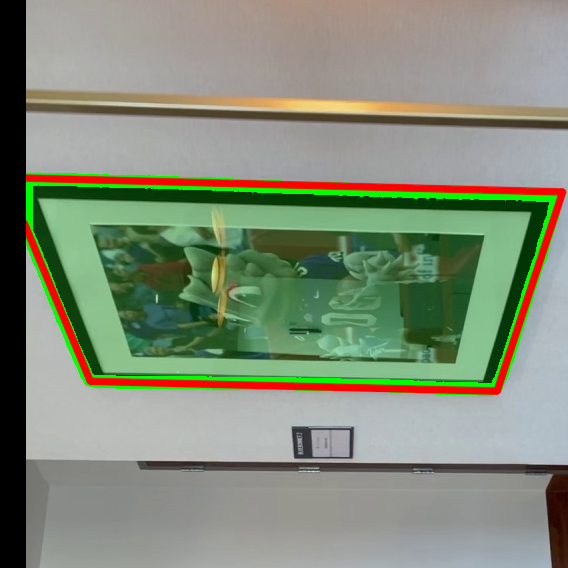}
    % (copy-file "/ssh:marr:/mnt/tlabdata/serycjon/flatsam/figures/Seq_00349_450.jpg" "~/dev/phd/doc/papers/cvpr2026/figures/sam/" t)
    \includegraphics[width=1\linewidth]{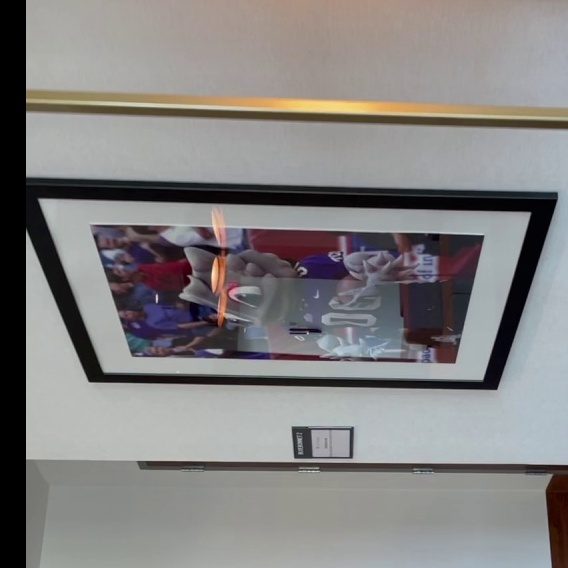}
    \caption{Reflective}
    \label{fig:sam-reflective}
  \end{subfigure}
  \begin{subfigure}[t]{0.135\linewidth}
    % (copy-file "/ssh:marr:/mnt/tlabdata/serycjon/flatsam/figures/Seq_00030_0_gt_mask.png" "~/dev/phd/doc/papers/cvpr2026/figures/sam/" t)
    \includegraphics[width=1\linewidth]{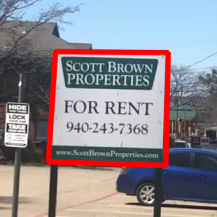}
    % (copy-file "/ssh:marr:/mnt/tlabdata/serycjon/flatsam/figures/Seq_00030_360_mask.png" "~/dev/phd/doc/papers/cvpr2026/figures/sam/" t)
    \includegraphics[width=1\linewidth]{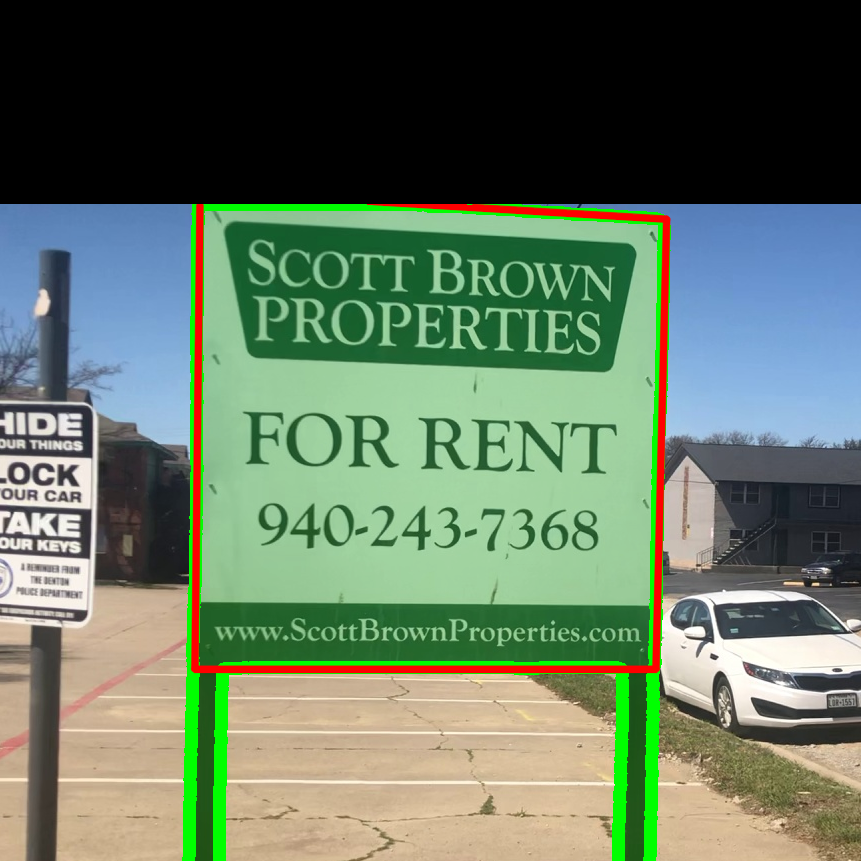}
    % (copy-file "/ssh:marr:/mnt/tlabdata/serycjon/flatsam/figures/Seq_00030_360.jpg" "~/dev/phd/doc/papers/cvpr2026/figures/sam/" t)
    \includegraphics[width=1\linewidth]{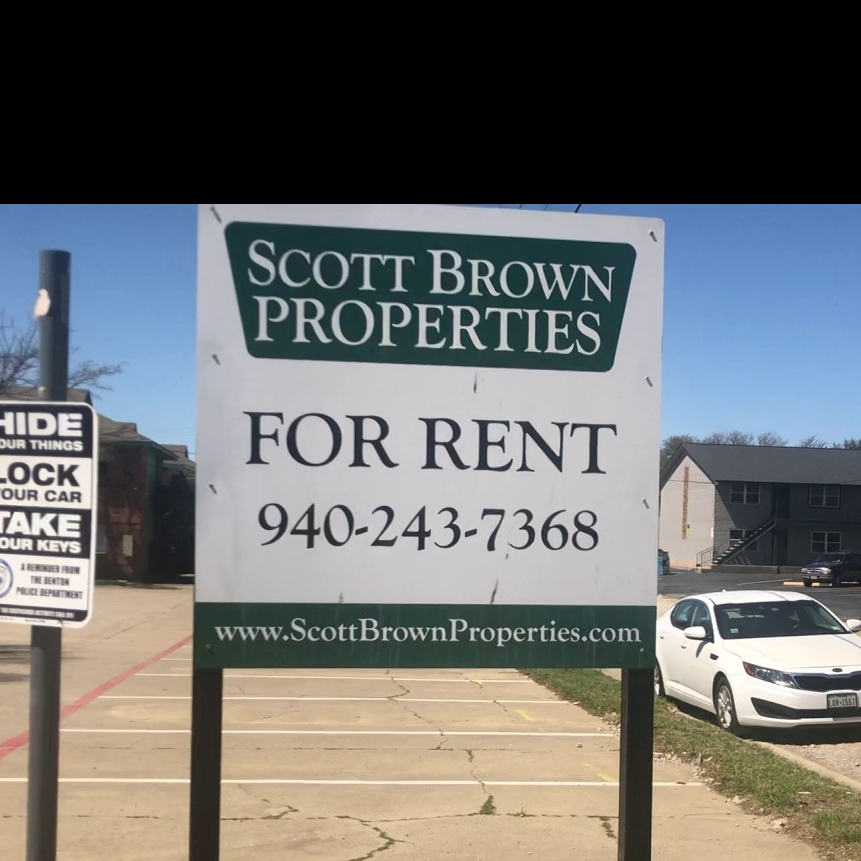}
    \caption{Spill}
    \label{fig:sam-spill-sign}
  \end{subfigure}
  \begin{subfigure}[t]{0.135\linewidth}
    % (copy-file "/ssh:marr:/mnt/tlabdata/serycjon/flatsam/figures/Seq_00006_0_gt_mask.png" "~/dev/phd/doc/papers/cvpr2026/figures/sam/" t)
    \includegraphics[width=1\linewidth]{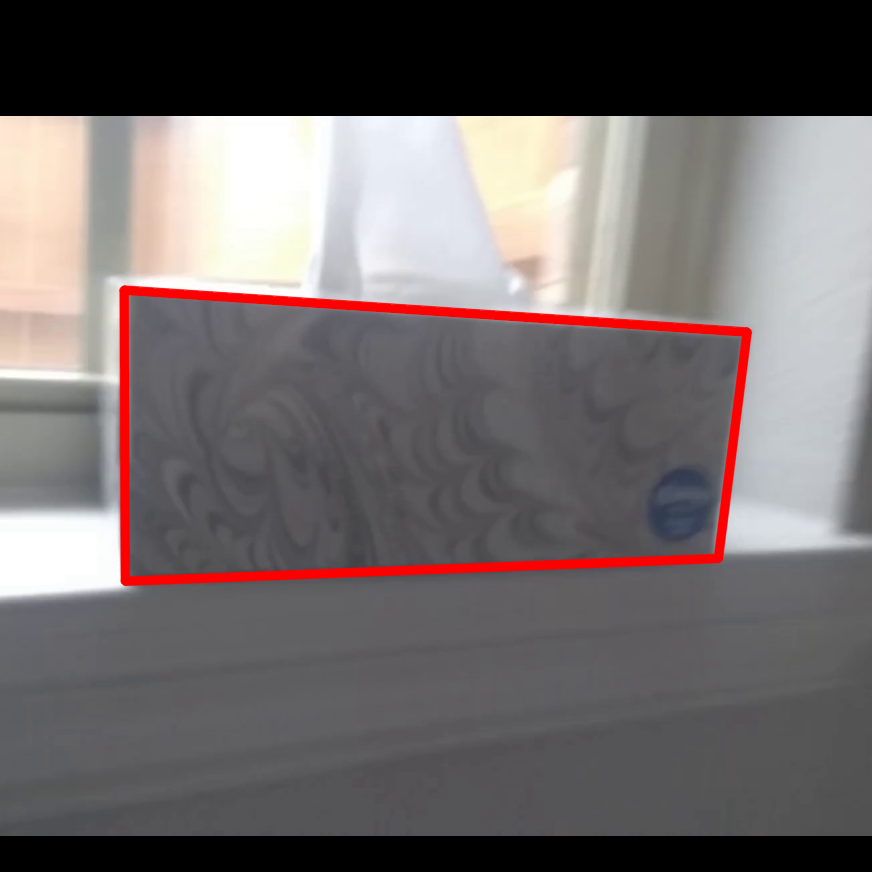}
    % (copy-file "/ssh:marr:/mnt/tlabdata/serycjon/flatsam/figures/Seq_00006_366_mask.png" "~/dev/phd/doc/papers/cvpr2026/figures/sam/" t)
    \includegraphics[width=1\linewidth]{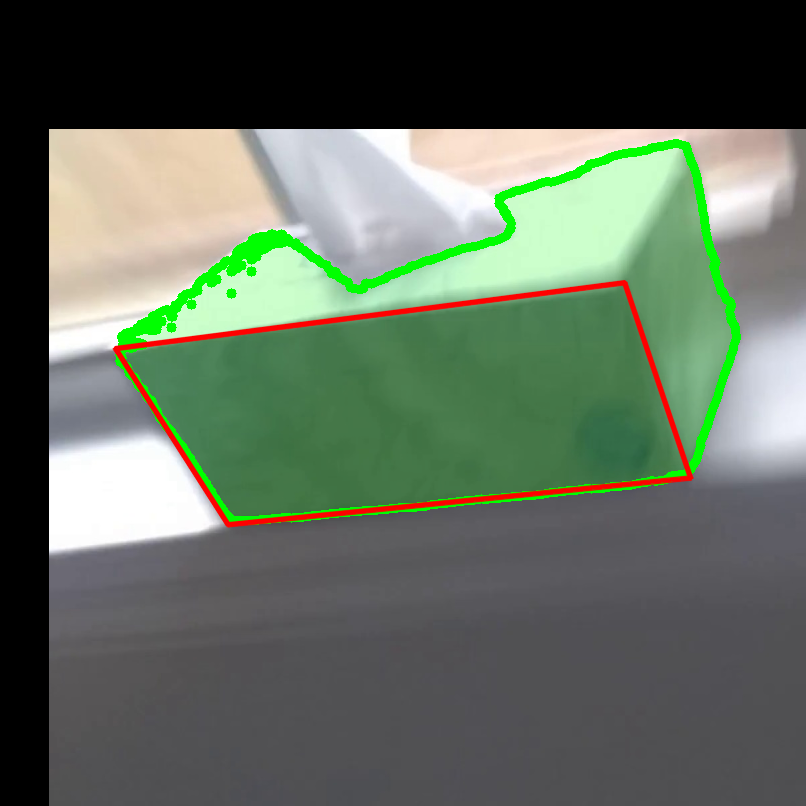}
    % (copy-file "/ssh:marr:/mnt/tlabdata/serycjon/flatsam/figures/Seq_00006_366.jpg" "~/dev/phd/doc/papers/cvpr2026/figures/sam/" t)
    \includegraphics[width=1\linewidth]{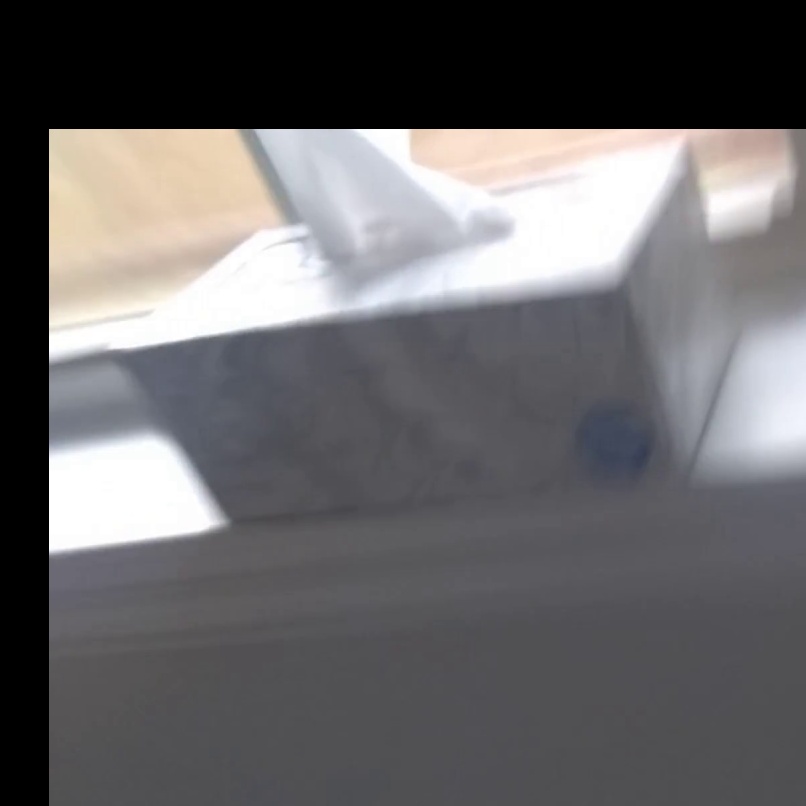}
    \caption{Box spill}
    \label{fig:sam-spill-box}
  \end{subfigure}
  \begin{subfigure}[t]{0.135\linewidth}
    % (copy-file "/ssh:marr:/mnt/tlabdata/serycjon/flatsam/figures/Seq_00305_0_gt_mask.png" "~/dev/phd/doc/papers/cvpr2026/figures/sam/" t)
    \includegraphics[width=1\linewidth]{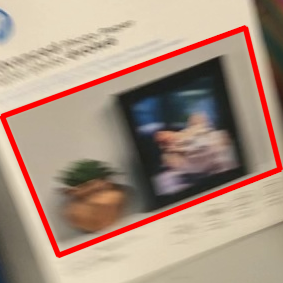}
    % (copy-file "/ssh:marr:/mnt/tlabdata/serycjon/flatsam/figures/Seq_00305_429_mask.png" "~/dev/phd/doc/papers/cvpr2026/figures/sam/" t)
    \includegraphics[width=1\linewidth]{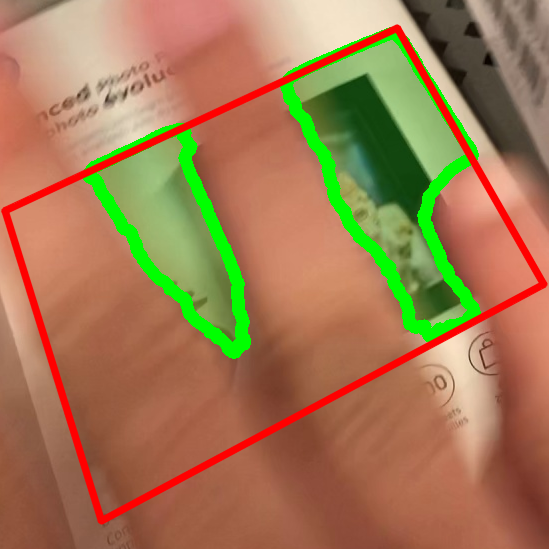}
    % (copy-file "/ssh:marr:/mnt/tlabdata/serycjon/flatsam/figures/Seq_00305_429.jpg" "~/dev/phd/doc/papers/cvpr2026/figures/sam/" t)
    \includegraphics[width=1\linewidth]{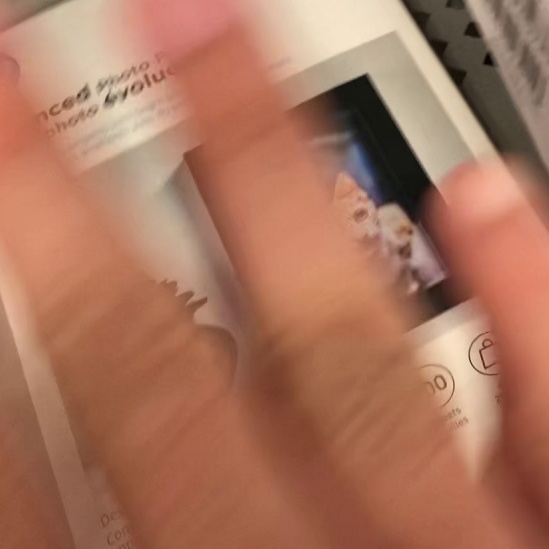}
    \caption{Occlusion}
    \label{fig:sam-occlusion}
  \end{subfigure}
  \begin{subfigure}[t]{0.135\linewidth}
    % (copy-file "/ssh:marr:/mnt/tlabdata/serycjon/flatsam/figures/Seq_00533_0_gt_mask.png" "~/dev/phd/doc/papers/cvpr2026/figures/sam/" t)
    \includegraphics[width=1\linewidth]{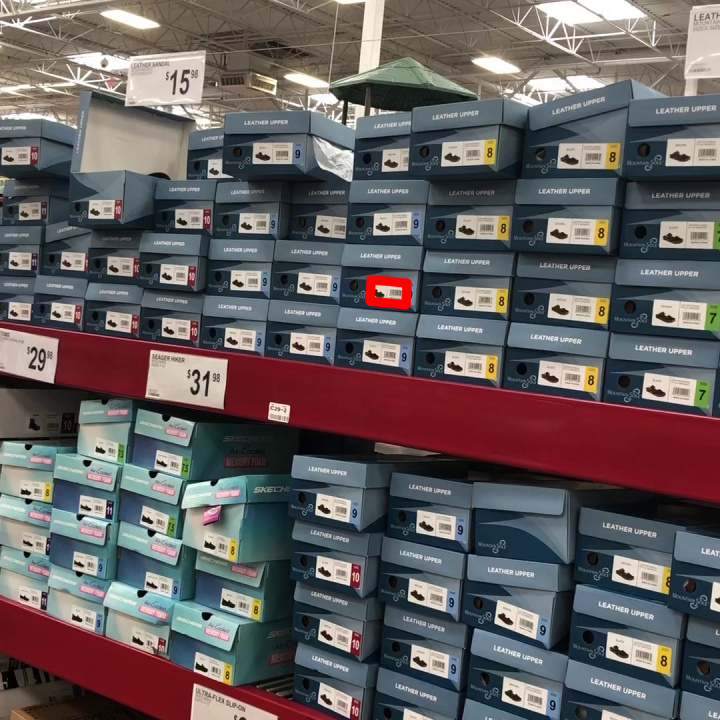}
    % (copy-file "/ssh:marr:/mnt/tlabdata/serycjon/flatsam/figures/Seq_00533_355_mask.png" "~/dev/phd/doc/papers/cvpr2026/figures/sam/" t)
    \includegraphics[width=1\linewidth]{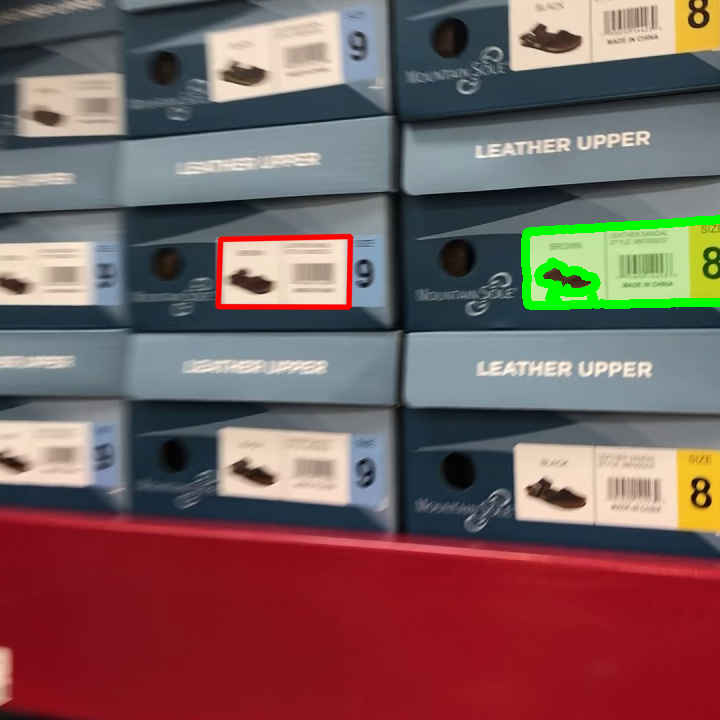}
    % (copy-file "/ssh:marr:/mnt/tlabdata/serycjon/flatsam/figures/Seq_00533_355.jpg" "~/dev/phd/doc/papers/cvpr2026/figures/sam/" t)
    \includegraphics[width=1\linewidth]{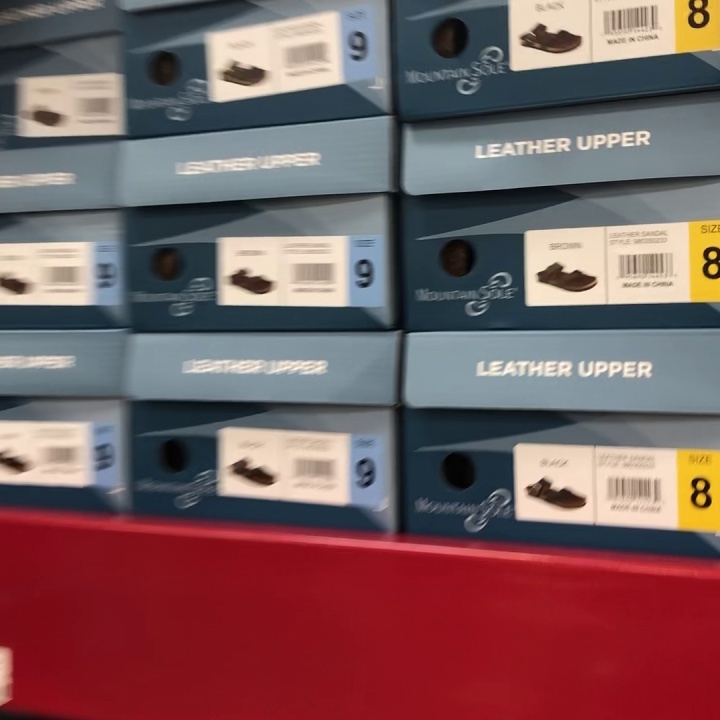}
    \caption{Distractor}
    \label{fig:sam-distractor}
  \end{subfigure}
  \caption{Examples of SAM 2~\cite{ravi2024sam2} output on selected PlanarTrack~\cite{liu2023planartrack} sequences.
    The images were cropped around the target from the original full-size frames.
    The initial frames shown on \emph{top} with  the ground truth pose in \emph{red}. 
    The \emph{middle row} shows the selected frame with SAM 2 segmentation overlays in \emph{green}.
    The \emph{bottom row} shows the frame without overlays for better idea of the input data.
    The first three examples, (\cref{fig:sam-blur,fig:sam-tless,fig:sam-reflective}), are difficult to track with correspondence-based methods, but are segmented well by SAM 2.
    In~\cref{fig:sam-spill-sign}, the segmentation ``spills'' outside the intended target, but gets corrected in the SAM-H Hough line detection step.
    Similar situation, but not recoverable by SAM-H is shown in~\cref{fig:sam-spill-box}.
    Segmentation works well during partial occlusions, but recovering the full pose based only on the mask is not always possible (\cref{fig:sam-occlusion}).
    Distractors sometimes cause SAM 2 tracking failure (\cref{fig:sam-distractor}).
  }
  \label{fig:sam-examples}
\end{figure*}

%% file: supplementary.tex
\clearpage
% Standalone/ESM build starts its own page numbering; in a combined build
% (arXiv: paper + supplementary in one PDF) the numbering must stay continuous.
\ifcombined\else\setcounter{page}{1}\fi
% \maketitlesupplementary
% give the supplementary its own hyperref destination namespace so the
% section counter reset below doesn't collide with the main paper's anchors
% (avoids wrong \Cref jumps into supplementary sections)
\renewcommand{\theHsection}{supp.\arabic{section}}
\setcounter{section}{0}
\renewcommand{\thesection}{\Alph{section}}
\renewcommand{\thesubsection}{\thesection.\arabic{subsection}}
\renewcommand{\thefigure}{S\arabic{figure}}
\setcounter{figure}{0}
\renewcommand{\thetable}{S\arabic{table}}
\setcounter{table}{0}

\begin{center}
{\Large \bf Segmentation-Guided Homography Estimation \\for Long-Term Planar Tracking}\\
\vspace{8mm}
{\large Supplementary Material}
\end{center}

\begin{figure*}
  \centering
  \setlength{\fboxsep}{0pt}
  % (copy-file "~/dev/phd/workarea/flatsam/fig_POT_success_plots_reannot.pdf" "~/dev/phd/doc/papers/cvpr2026/figures/" t)
  % \framebox{\includegraphics[page=1,width=0.32\linewidth]{fig_POT_success_plots_reannot}} % 
  \fbox{\includegraphics[page=1,width=0.32\linewidth]{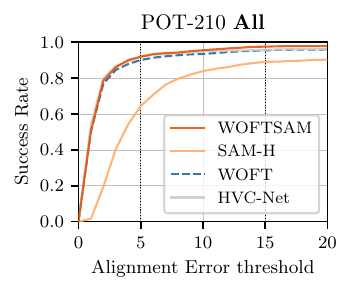}} % 
  \includegraphics[page=2,width=0.32\linewidth]{fig_POT_success_plots_reannot} %
  \includegraphics[page=3,width=0.32\linewidth]{fig_POT_success_plots_reannot} \\
  \includegraphics[page=4,width=0.32\linewidth]{fig_POT_success_plots_reannot} %
  \includegraphics[page=5,width=0.32\linewidth]{fig_POT_success_plots_reannot} %
  \includegraphics[page=6,width=0.32\linewidth]{fig_POT_success_plots_reannot} \\
  \includegraphics[page=7,width=0.32\linewidth]{fig_POT_success_plots_reannot} %
  \includegraphics[page=8,width=0.32\linewidth]{fig_POT_success_plots_reannot} %
  % \framebox{\includegraphics[page=1,width=0.32\linewidth]{POT210_merged_AE_roc}} %
  \caption{Alignment Error success plots on POT-210~\cite{liang2017planar}, with precise GT re-annotation from~\cite{serych2023planar}.
    The results are shown on the whole dataset (\textbf{All}) and also split by the POT-210 challenge types.
  Thanks to the proposed \samh{}-based re-detection scheme, WOFTSAM performs better than the WOFT baseline.  The p@5 and p@15 metric thresholds are highlighted with dotted lines.}
  \label{fig:pot-success-plots}
\end{figure*}
\section{Detailed POT-210 Results}
\label{sec:more-pot-results} 
\begin{figure*}
  \centering
  % parallel -j 4 python experiments/paper_sam_vis.py /mnt/tlabdata/serycjon/flatsam/{pot_woftsam_samfallback_nl,POT-SAM-H,pot_woft_replica}/ V30_5 {} --thickness 4 -ar 1 --crop_margin 0.2 --dataset pot --bw -w --out_dir /mnt/tlabdata/serycjon/flatsam/figures_supp ::: 341 343 345 347
  % (copy-file "~/dev/phd/data/export/flatsam/figures_supp/V30_5_0_gt_mask.png" "~/dev/phd/doc/papers/cvpr2026/figures/pot-tracking-examples/" t)
  % (copy-file "~/dev/phd/data/export/flatsam/figures_supp/V30_5_341_mask.png" "~/dev/phd/doc/papers/cvpr2026/figures/pot-tracking-examples/" t)
  % (copy-file "~/dev/phd/data/export/flatsam/figures_supp/V30_5_343_mask.png" "~/dev/phd/doc/papers/cvpr2026/figures/pot-tracking-examples/" t)
  % (copy-file "~/dev/phd/data/export/flatsam/figures_supp/V30_5_345_mask.png" "~/dev/phd/doc/papers/cvpr2026/figures/pot-tracking-examples/" t)
  % (copy-file "~/dev/phd/data/export/flatsam/figures_supp/V30_5_347_mask.png" "~/dev/phd/doc/papers/cvpr2026/figures/pot-tracking-examples/" t)
  \includegraphics[width=0.20\linewidth]{pot-tracking-examples/V30_5_0_gt_mask}%
  \includegraphics[width=0.20\linewidth]{pot-tracking-examples/V30_5_341_mask}%
  \includegraphics[width=0.20\linewidth]{pot-tracking-examples/V30_5_343_mask}%
  \includegraphics[width=0.20\linewidth]{pot-tracking-examples/V30_5_345_mask}%
  \includegraphics[width=0.20\linewidth]{pot-tracking-examples/V30_5_347_mask}\\
  % parallel -j 4 python experiments/paper_sam_vis.py /mnt/tlabdata/serycjon/flatsam/{pot_woftsam_samfallback_nl,POT-SAM-H,pot_woft_replica}/ V10_5 {} --thickness 4 -ar 1 --crop_margin 0.2 --dataset pot --bw -w --out_dir /mnt/tlabdata/serycjon/flatsam/figures_supp ::: 65 77 137 176
  % (copy-file "~/dev/phd/data/export/flatsam/figures_supp/V10_5_0_gt_mask.png" "~/dev/phd/doc/papers/cvpr2026/figures/pot-tracking-examples/" t)
  % (copy-file "~/dev/phd/data/export/flatsam/figures_supp/V10_5_65_mask.png" "~/dev/phd/doc/papers/cvpr2026/figures/pot-tracking-examples/" t)
  % (copy-file "~/dev/phd/data/export/flatsam/figures_supp/V10_5_77_mask.png" "~/dev/phd/doc/papers/cvpr2026/figures/pot-tracking-examples/" t)
  % (copy-file "~/dev/phd/data/export/flatsam/figures_supp/V10_5_137_mask.png" "~/dev/phd/doc/papers/cvpr2026/figures/pot-tracking-examples/" t)
  % (copy-file "~/dev/phd/data/export/flatsam/figures_supp/V10_5_176_mask.png" "~/dev/phd/doc/papers/cvpr2026/figures/pot-tracking-examples/" t)
  \includegraphics[width=0.20\linewidth]{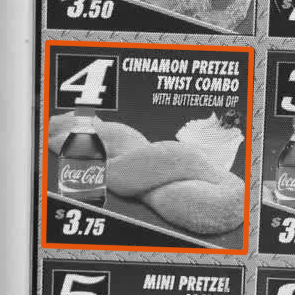}% 
  \includegraphics[width=0.20\linewidth]{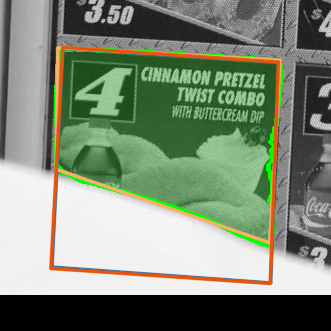}%
  \includegraphics[width=0.20\linewidth]{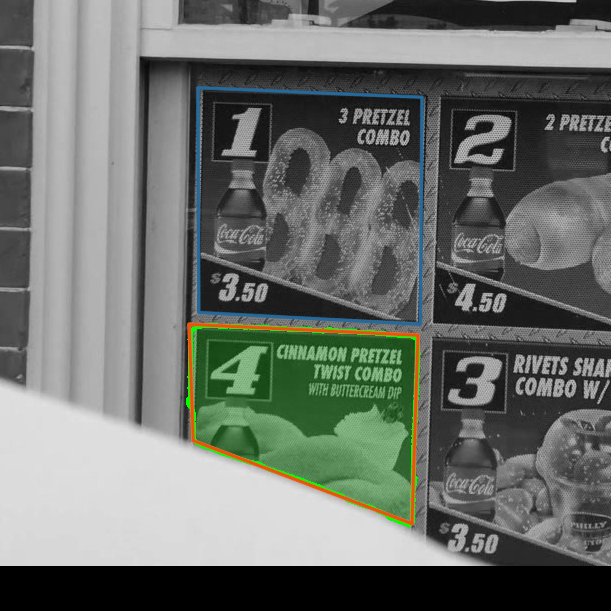}%
  \includegraphics[width=0.20\linewidth]{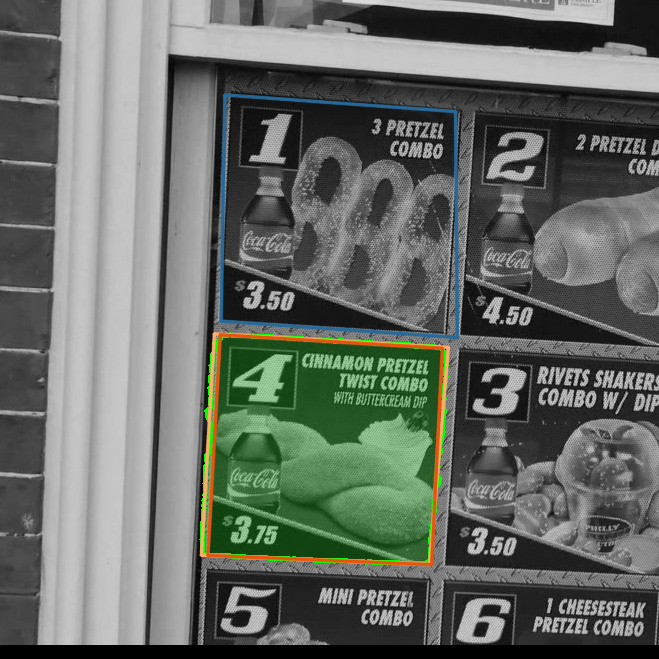}%
  \includegraphics[width=0.20\linewidth]{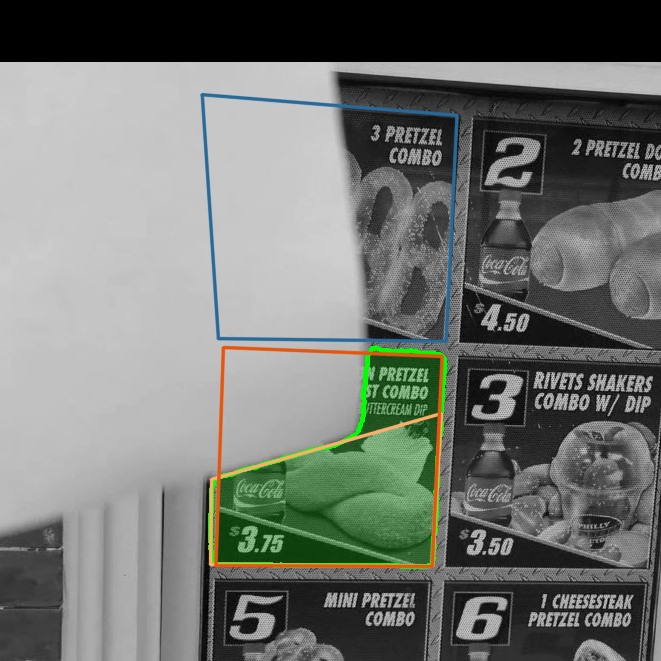}
  \caption{POT-210~\cite{liang2017planar} examples, where outputs of WOFTSAM (\textcolor{woftsamcolor}{\emph{dark orange}}) significantly differ from the outputs of \samh{} (\textcolor{samhcolor}{\emph{light orange}}) or WOFT (\textcolor{woftcolor}{\emph{blue}}).
    SAM 2 mask shown as a \textcolor{green!80!black}{\emph{green}} overlay, frames are cropped around the object.\\
    \emph{Top:} \samh{} homography is incorrect due to occlusion with a linear boundary.
    WOFT and WOFTSAM outputs overlap.
    Frames $I_0, I_{341}, I_{343}, I_{345}, I_{347}$ of \texttt{V30\_5}.\\
    \emph{Bottom:} WOFT switches to tracking an incorrect target during occlusion (\emph{middle image}).
    \samh{} stays on the correct object, although with an incorrect homography during the occlusion.
    WOFTSAM stays on the correct target thanks to the robust re-detection (\emph{last image}).
    WOFT and WOFTSAM outputs overlap on the first two images.
    Frames $I_0, I_{65}, I_{77}, I_{137}, I_{176}$ of \texttt{V10\_5}.
  }
  \label{fig:pot-tracking-examples}
\end{figure*}

We provide the commonly reported alignment error success plots in~\Cref{fig:pot-success-plots}.
WOFTSAM consistently out-performs the baseline WOFT~\cite{serych2023planar}.
At the 15\,px alignment error threshold, the failure rate is almost halved.
The improvement comes mainly from the \emph{blur}, \emph{occlusion}, and \emph{unconstrained} sequences, where the re-detection ability is most needed.
Robust re-detection also helps in some of the \emph{scale} sequences, where the baseline WOFT fails to keep track of a small (zoomed out) object.
Some examples of the WOFTSAM, \samh{} and WOFT outputs are shown in~\Cref{fig:pot-tracking-examples}.
% We also attach the raw tracking results in the \texttt{POT210Results.tgz} archive.

% \section{SAM Output Examples on PlanarTrack}
% \label{sec:sam-output-examples}
% In~\Cref{fig:sam-examples} we show example outputs of the SAM 2~\cite{ravi2024sam2} segmentation tracker, demonstrating some of its strengths and failure modes.

% \section{Detailed PlanarTrack Results}
% \label{sec:more-pt-results}
% Similar to POT-210, we show the success rate curves for the PlanarTrack~\cite{liu2023planartrack} dataset in~\Cref{fig:planartrack-success-plots}.
% The plots show the proposed WOFTSAM to be consistently better than WOFT~\cite{serych2023planar} on all error thresholds.

% Note that in contrast to POT-210 (where each video is of one type, \eg \emph{occlusion}, or \emph{scale}), each video is assigned multiple challenging attributes in PlanarTrack.
% For example, 271 of the 300 sequences have four or more attributes assigned (out of the total of eight), and 114 sequences have six or more attributes.
% Consequently, the per-attribute results are not easily interpretable.
% Most common combinations of attributes are shown in~\Cref{fig:pt-upset}.
% In general, the PlanarTrack sequences would be categorized as \emph{unconstrained} in the POT-210 terminology.

\section{Re-Annotation Details}
\label{sec:re-annot-details}
The annotation tool is controlled by keyboard, allowing the annotator to select a control point to move and moving it up, down, left, or right in increments of $N$\,px.
The movement step is set to $N=1$ by default, but can be repeatedly multiplied or divided by 2 to correct larger errors and fine-tune precisely.
Apart for a zoomed-in view of the initial frame and the control-point-defined quadrilateral, the annotation tool also shows a later frame for reference.
By default, the frame with the largest target object area is shown, but the video can be stepped in both directions to show a different reference frame.
This is to ensure that we adhere precisely to what the original annotators were annotating.
Moreover, a third view shows image intensity alignment between the current initial frame annotation and the selected reference frame GT.
This way, the annotator can find a precisely annotated reference frame and manually align the initial frame with it, to achieve sub-pixel precision.
Examples of the resulting re-annotation quality are shown in~\Cref{fig:reannot-more-examples}, in addition to~\xref{\Cref{fig:pt-init-reannot}}{Figure~9} in the paper.

Error statistics of the original GT annotations compared to our re-annotation are shown in~\Cref{fig:reannot-stats}.
Note how relatively small alignment error of the first frame annotation significantly affects the evaluation results (\xref{\Cref{tab:main-results}}{Table~1} in the paper). The annotation tool will be made publicly available together with the WOFTSAM and \samh{} code.
% at~\url{https://github.com/serycjon/WOFTSAM}.
\begin{figure*}
  \centering
  \newcommand{\annotsize}{0.32}
  \input{figures/reannot/PT-init-reannotation-vis-Seq_00046.tex}
  \input{figures/reannot/PT-init-reannotation-vis-Seq_00023.tex}
  \input{figures/reannot/PT-init-reannotation-vis-Seq_00403.tex}\\
  \input{figures/reannot/PT-init-reannotation-vis-Seq_00060.tex}
  \input{figures/reannot/PT-init-reannotation-vis-Seq_00250.tex}
  \input{figures/reannot/PT-init-reannotation-vis-Seq_00271.tex}\\
  \input{figures/reannot/PT-init-reannotation-vis-Seq_00161.tex}
  \input{figures/reannot/PT-init-reannotation-vis-Seq_00040.tex}
  \input{figures/reannot/PT-init-reannotation-vis-Seq_00153.tex}
  \caption{Examples of our PlanarTrack\textsubscript{TST}~\cite{liu2023planartrack} re-annotation on the initial frames.
  The original GT (\emph{red}) does not copy the object boundaries precisely, while the new improved GT (\emph{cyan}) aligns with the target.}
  \label{fig:reannot-more-examples}
\end{figure*}
\pagebreak
\section{WOFTSAM efficiency}
\label{sec:speed}
\begin{wrapfigure}{r}{0.4\textwidth}
  \centering
  % (copy-file "~/dev/phd/workarea/flatsam/reannotation_ae_hist.pdf" "~/dev/phd/doc/papers/cvpr2026/figures/" t)
  \includegraphics[width=1\linewidth]{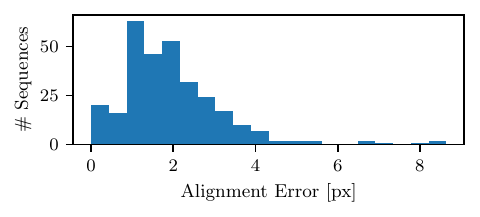} \\
  % (copy-file "~/dev/phd/workarea/flatsam/reannotation_dist_hist.pdf" "~/dev/phd/doc/papers/cvpr2026/figures/" t)
  \includegraphics[width=1\linewidth]{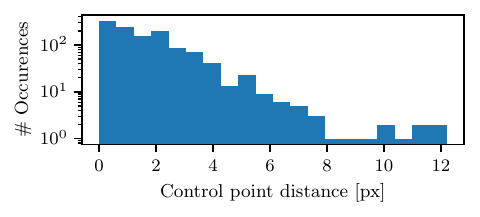}
  \caption{Error histograms of the original PlanarTrack\textsubscript{TST}~\cite{liu2023planartrack} initial frame annotations compared to our re-annotation.
    \emph{Top:} alignment error $e_\text{AL}$, capturing the annotation difference over all the four control points (RMSE).
    \emph{Bottom:} euclidean distance of individual control point positions, with y-axis scaled logarithmically.
  }
  \label{fig:reannot-stats}
\end{wrapfigure}
We evaluated the efficiency on the PlanarTrack\textsubscript{TST} dataset. Overall, including loading the \(1280 \times 720\) images from disk, the proposed WOFTSAM runs at 422\,ms/frame (2.4\,FPS), using a NVIDIA RTX A5000 GPU, requiring 4.5\,GB of GPU RAM (and Intel(R) Xeon(R) Gold 6326 CPU).
On the same device, WOFT runs at a similar speed of 2.7\,FPS.

Here we provide a detailed analysis of the runtime of the most heavy parts of the computation.
The SAM2 segmentation takes only 42\,ms per frame (using the hiera tiny SAM2.1 model), which could be further reduced or even removed by computing the segmentation in parallel with the first pre-warp + Weighted-Flow-Homography (part 1 in~\xref{\Cref{fig:woftsam_overview}}{Figure~3} in the paper).
The DINOv2-based symmetry resolution takes 102\,ms on average, but is only used in around 5\% of frames, resulting in an amortized 5\,ms runtime.
A single RAFT optical flow computation takes 170\,ms on average.
The Hough line detection and post-processing takes 132\,ms on average, which would be significantly improved by rewriting from Python to e.g. C/C++.
The most computationally costly part is the optical flow, just like in WOFT.
While the WOFTSAM doesn't currently run real-time, it is usable for interactive applications like tracking in movie post-production.

\section{Hough Transform Details}
\label{sec:more-hough}

This section describes the implementation details of the Hough transform for line detection.
First, we discard all mask contours shorter than 20\,px (SAM noise) and normalize the rest to have zero mean, moving the center of the target into origin.
Then we do the Hough voting followed by least squares refinement.
The Hough accumulator represents the lines with normal angle quantized to 359 values, \ie, 1° step, and 100 distances from origin, spanning from 0 to maximum distance of a contour point.
We traverse the contours one-by-one.
For each contour point $x_i$, we compute the contour direction as the angle $\alpha_i$ between $x_{i-4}, x_{i+4}$ (indexes modulo contour length).
Each point casts multiple votes for lines going through it.
In particular we consider lines differing from the angle $\alpha_i$ by $\Delta\alpha \in \{-10^\circ, -8^\circ, \dots, 0^\circ, \dots, +10^\circ\}$ (11 votes).
We ensure that distances are non-negative by rotating $180^\circ$ when needed.
The votes are weighted by $w = 1 - \frac{|\Delta\alpha|}{20}$, assigning full weight 1 to the original $\alpha_i$, half the weight 0.5 to the extremes $\alpha_i \pm 10^\circ$, and linearly ramping in between.

Next we find peaks in the accumulator.
We repeat the accumulator 3 times side by side (unwrapping the periodic angle axis), and perform a Gaussian filtering with $\sigma = 4$.
To find peaks in the resulting smoothed accumulator, we use \texttt{find\_peaks} from SciPy, with minimal peak distance of 10.
After undoing the accumulator unwrap, \ie, discarding the peaks from its first and its last repetition, only keeping the peaks in the middle, we take the top $K=4$ ones, sort them by the angle and convert back to line equations in the original image coordinates.

This procedure is followed by a refinement step.
The contours are traversed once again, computing the direction $\alpha_i$ between $x_{i-4}, x_{i+4}$ as before.
The contour point $x_i$ is assigned to the closest one of the four Hough-detected lines $l_j$, if it is close enough, $\text{d}(x_i, l_j) \le 15\,\text{px}$ and the direction $\alpha_i$ is less than 15° from the Hough line direction $\beta_j$.
After collecting these support points $S_j = \{x_i \mid i = \argmin_i \text{dist}(x_i, l_j); \text{d}(x_i, l_j) \le 15\,\text{px}; \text{d}(\alpha_i, \beta_j) \le 15^\circ\}$ for each line, the lines get re-computed by a least-squares fit through their supporting points $S_j$.

The results of \samh{} are stable across a range of hyperparameters as shown in~\cref{tab:hough_hyperparams}.
Most significant performance drop using too large Hough accumulator smoothing \(\sigma\) (-1.9pp p@15).
\begin{table}
  \centering
  \begin{tabular}{cccc|ll}
    \toprule
    voting window \(\pm ^\circ\) & smoothing \(\sigma\) & line-point px & line-point \(^\circ\) & PT\textsubscript{TST} p@5        & PT\textsubscript{TST} p@15       \\
    % A                              & B                    & C             & D                     & PT p@5        & PT p@15       \\
    \midrule
    10                             & 4                    & 15            & 15                    & 62.4          & 80.5          \\
    \midrule
    5                              &                      &               &                       & 62.4 \tiny{(\(+0.0\))} & 80.5 \tiny{(\(-0.0\))} \\
    20                             &                      &               &                       & 61.9 \tiny{(\(-0.5\))} & 79.8 \tiny{(\(-0.7\))} \\
                                   & 2                    &               &                       & 62.2 \tiny{(\(-0.2\))} & 80.1 \tiny{(\(-0.4\))} \\
                                   & 8                    &               &                       & 60.8 \tiny{(\(-1.6\))} & 78.6 \tiny{(\(-1.9\))} \\
                                   &                      & 10            &                       & 62.5 \tiny{(\(+0.1\))} & 80.5 \tiny{(\(+0.0\))} \\
                                   &                      & 25            &                       & 62.3 \tiny{(\(-0.1\))} & 80.6 \tiny{(\(+0.0\))} \\
                                   &                      &               & 10                    & 62.3 \tiny{(\(-0.1\))} & 80.5 \tiny{(\(-0.1\))} \\
                                   &                      &               & 25                    & 62.2 \tiny{(\(-0.2\))} & 80.1 \tiny{(\(-0.4\))} \\
    \bottomrule
  \end{tabular}
  \caption{Hough hyperparameters evaluated on the PlanarTrack\textsubscript{TST} dataset with re-annotated init frames -- \samh{} is stable in a wide range.
    % A: voting window angle range, B: smoothing \(\sigma\), C: line-point distance threshold, D: line-point angle threshold.
    First row is the default \samh{}, other rows alter one hyperparameter. 
    Computed on newer stack than the main paper results (NVIDIA A100 GPU, AMD EPYC 7543 CPU, CUDA 12.6, CuDNN 9.5.0, PyTorch 2.7.1), leading to a small difference in the results (+0.4pp on both PT thresholds for \samh{}). The method itself did not change.
  }
  \label{tab:hough_hyperparams}
\end{table}

\section{Baseline Experiments}
\label{sec:baseline-experiments}
We provide additional baseline experiment results in~\cref{tab:baselines}.
First, we show that \samh{} with no DINO-based symmetry disambiguation, \ie relying only on the small-motion criterion, performs significantly worse than the full \samh{} (-12.5pp PlanarTrack p@15).

Second, we evaluated two mask-to-quadrilateral baselines as replacements of the \samh{} method: min-area rectangle (rotated bounding-box) and Ramer-Douglas-Peucker (RDP) approximation of the segmentation mask contour (the RDP parameter was optimized on each frame so that the result is a quadrilateral).
Both of these mask-to-quad baselines use motion-only symmetry disambiguation (like the ``\samh{} no DINO'' row).

\begin{table}
  \centering
  \begin{tabular}{lll}
    \toprule
    method             & p@5         & p@15        \\
    \midrule
    \samh{}              & 62.4 \tiny{(\(+8.7\))}  & 80.5 \tiny{(\(+12.5\))} \\
    \samh{} no DINO      & 53.7                    & 68.0                    \\
    min-area rectangle & 13.6 \tiny{(\(-40.0\))} & 39.2 \tiny{(\(-28.8\))} \\
    RDP approximation  & 39.1 \tiny{(\(-14.5\))} & 70.1 \tiny{(\(+2.1\))}  \\
    \bottomrule
  \end{tabular}
  \caption{Symmetry disambiguation and mask-to-quad baselines evaluated on the PlanarTrack\textsubscript{TST} dataset with re-annotated init frames.
    The proposed \samh{} achieves significantly better performance than the baselines.
    Computed on newer stack than the main paper results (NVIDIA A100 GPU, AMD EPYC 7543 CPU, CUDA 12.6, CuDNN 9.5.0, PyTorch 2.7.1), leading to a small difference in the results (+0.4pp on both PT thresholds for \samh{}). The method itself did not change.
  }
  \label{tab:baselines}
\end{table}

%% file: figures/reannot/PT-init-reannotation-vis-Seq_00046.tex
\resizebox{\annotsize\linewidth}{!}{%
            \begin{tikzpicture}[x=1pt, y=1pt]
            \node[anchor=south west, inner sep=0] at (0,0) {\includegraphics{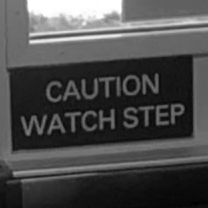}};\definecolor{tmpcolor}{rgb}{255,0,0}
\draw[tmpcolor, thick]
(8.741264999999999, 138.73627299999998) -- (195.62978399999997, 153.112313) -- (198.02579000000003, 72.84609) -- (11.137271000000055, 59.668052999999986) -- cycle;
\definecolor{tmpcolor}{rgb}{0,255,255}
\draw[tmpcolor, thick]
(8.991300000000024, 138.73630000000003) -- (192.87980000000005, 153.1123) -- (193.0258, 69.84609999999998) -- (12.387299999999982, 56.41809999999998) -- cycle;
\end{tikzpicture}}

%% file: figures/reannot/PT-init-reannotation-vis-Seq_00023.tex
\resizebox{\annotsize\linewidth}{!}{%
            \begin{tikzpicture}[x=1pt, y=1pt]
            \node[anchor=south west, inner sep=0] at (0,0) {\includegraphics{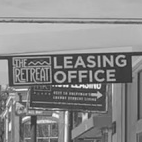}};\definecolor{tmpcolor}{rgb}{255,0,0}
\draw[tmpcolor, thick]
(6.4567389999999705, 83.97421000000003) -- (132.24708799999996, 89.964226) -- (134.64309500000002, 57.61813599999999) -- (7.654742000000056, 52.826122999999995) -- cycle;
\definecolor{tmpcolor}{rgb}{0,255,255}
\draw[tmpcolor, thick]
(7.456699999999955, 86.9742) -- (130.49710000000005, 90.2142) -- (131.8931, 59.36810000000003) -- (8.654700000000048, 55.3261) -- cycle;
\end{tikzpicture}}

%% file: figures/reannot/PT-init-reannotation-vis-Seq_00403.tex
\resizebox{\annotsize\linewidth}{!}{%
            \begin{tikzpicture}[x=1pt, y=1pt]
            \node[anchor=south west, inner sep=0] at (0,0) {\includegraphics{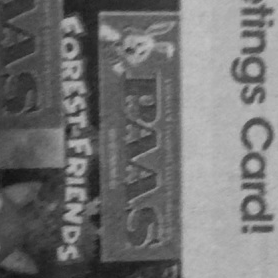}};\definecolor{tmpcolor}{rgb}{255,0,0}
\draw[tmpcolor, thick]
(99.57986699999998, 266.103993) -- (182.641431, 266.103993) -- (182.641431, 19.04908499999999) -- (95.32029999999997, 12.659733999999958) -- cycle;
\definecolor{tmpcolor}{rgb}{0,255,255}
\draw[tmpcolor, thick]
(98.57990000000001, 265.104) -- (181.64139999999998, 265.104) -- (181.64139999999998, 21.049099999999953) -- (101.32029999999997, 17.659700000000043) -- cycle;
\end{tikzpicture}}

%% file: figures/reannot/PT-init-reannotation-vis-Seq_00060.tex
\resizebox{\annotsize\linewidth}{!}{%
            \begin{tikzpicture}[x=1pt, y=1pt]
            \node[anchor=south west, inner sep=0] at (0,0) {\includegraphics{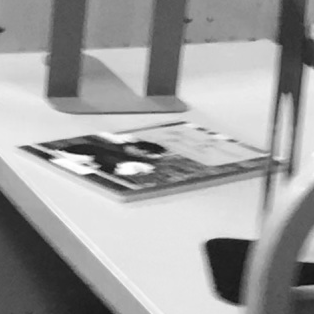}};\definecolor{tmpcolor}{rgb}{255,0,0}
\draw[tmpcolor, thick]
(176.66571899999997, 200.901818) -- (299.1148290000001, 150.32501200000002) -- (121.84968900000001, 114.06623300000001) -- (14.399515999999949, 170.12719299999998) -- cycle;
\definecolor{tmpcolor}{rgb}{0,255,255}
\draw[tmpcolor, thick]
(182.66570000000002, 196.90179999999998) -- (291.11480000000006, 154.325) -- (121.84969999999998, 121.06619999999998) -- (14.399499999999989, 168.12720000000002) -- cycle;
\end{tikzpicture}}

%% file: figures/reannot/PT-init-reannotation-vis-Seq_00250.tex
\resizebox{\annotsize\linewidth}{!}{%
            \begin{tikzpicture}[x=1pt, y=1pt]
            \node[anchor=south west, inner sep=0] at (0,0) {\includegraphics{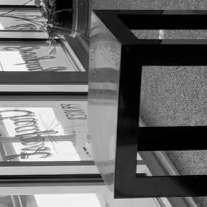}};\definecolor{tmpcolor}{rgb}{255,0,0}
\draw[tmpcolor, thick]
(88.122296, 195.96588999999994) -- (120.069052, 159.75956699999995) -- (115.809484, 10.674709000000007) -- (85.99251199999998, 51.14059899999995) -- cycle;
\definecolor{tmpcolor}{rgb}{0,255,255}
\draw[tmpcolor, thick]
(91.1223, 197.96590000000003) -- (120.06909999999999, 163.75959999999998) -- (112.80950000000001, 9.67470000000003) -- (85.9925, 59.14059999999995) -- cycle;
\end{tikzpicture}}

%% file: figures/reannot/PT-init-reannotation-vis-Seq_00271.tex
\resizebox{\annotsize\linewidth}{!}{%
            \begin{tikzpicture}[x=1pt, y=1pt]
            \node[anchor=south west, inner sep=0] at (0,0) {\includegraphics{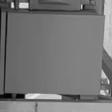}};\definecolor{tmpcolor}{rgb}{255,0,0}
\draw[tmpcolor, thick]
(9.877703999999994, 98.39351099999999) -- (105.71796999999998, 98.39351099999999) -- (101.45840299999998, 17.46172999999999) -- (5.618135999999993, 15.331947000000014) -- cycle;
\definecolor{tmpcolor}{rgb}{0,255,255}
\draw[tmpcolor, thick]
(7.877700000000004, 98.39350000000002) -- (104.71800000000002, 97.39350000000002) -- (99.45839999999998, 18.46169999999995) -- (4.618100000000027, 19.33190000000002) -- cycle;
\end{tikzpicture}}

%% file: figures/reannot/PT-init-reannotation-vis-Seq_00161.tex
\resizebox{\annotsize\linewidth}{!}{%
            \begin{tikzpicture}[x=1pt, y=1pt]
            \node[anchor=south west, inner sep=0] at (0,0) {\includegraphics{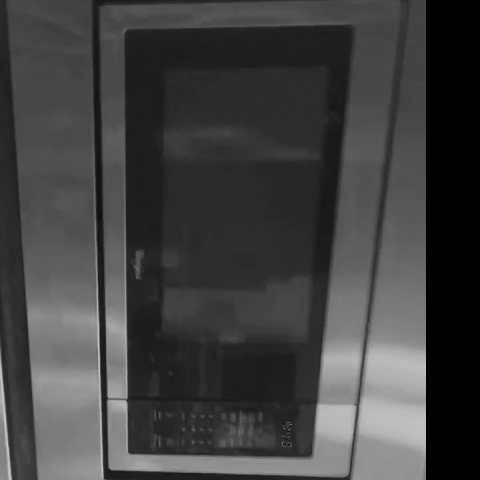}};\definecolor{tmpcolor}{rgb}{255,0,0}
\draw[tmpcolor, thick]
(129.328619, 450.403494) -- (352.955907, 458.92262900000003) -- (316.749584, 24.446755000000053) -- (125.069052, 22.316971999999964) -- cycle;
\definecolor{tmpcolor}{rgb}{0,255,255}
\draw[tmpcolor, thick]
(124.3286, 450.4035) -- (354.95590000000004, 457.9226) -- (312.7496, 22.446800000000053) -- (128.0691, 26.317000000000007) -- cycle;
\end{tikzpicture}}

%% file: figures/reannot/PT-init-reannotation-vis-Seq_00040.tex
\resizebox{\annotsize\linewidth}{!}{%
            \begin{tikzpicture}[x=1pt, y=1pt]
            \node[anchor=south west, inner sep=0] at (0,0) {\includegraphics{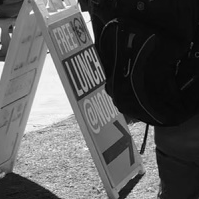}};\definecolor{tmpcolor}{rgb}{255,0,0}
\draw[tmpcolor, thick]
(48.97088199999996, 171.674709) -- (83.71297800000002, 189.644759) -- (149.603161, 36.300332999999995) -- (116.05906800000002, 11.142263000000014) -- cycle;
\definecolor{tmpcolor}{rgb}{0,255,255}
\draw[tmpcolor, thick]
(46.97090000000003, 173.4247) -- (79.96299999999997, 187.6448) -- (143.35320000000002, 36.80029999999999) -- (112.55909999999994, 10.392299999999977) -- cycle;
\end{tikzpicture}}

%% file: figures/reannot/PT-init-reannotation-vis-Seq_00153.tex
\resizebox{\annotsize\linewidth}{!}{%
            \begin{tikzpicture}[x=1pt, y=1pt]
            \node[anchor=south west, inner sep=0] at (0,0) {\includegraphics{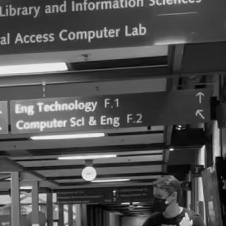}};\definecolor{tmpcolor}{rgb}{255,0,0}
\draw[tmpcolor, thick]
(9.999168000000054, 123.794509) -- (208.86771999999996, 134.576539) -- (214.85773700000004, 99.834443) -- (11.197171000000026, 91.448419) -- cycle;
\definecolor{tmpcolor}{rgb}{0,255,255}
\draw[tmpcolor, thick]
(9.999199999999973, 125.7945) -- (208.3677, 137.5765) -- (209.10770000000002, 103.33439999999999) -- (10.947199999999953, 92.44839999999999) -- cycle;
\end{tikzpicture}}